\documentclass[pdflatex,sn-aps]{sn-jnl}

\usepackage{graphicx}%
\usepackage{multirow}%
\usepackage{amsmath,amssymb,amsfonts}%
\usepackage{amsthm}%
\usepackage{mathrsfs}%
\usepackage[title]{appendix}%
\usepackage{xcolor}%
\usepackage{textcomp}%
\usepackage{manyfoot}%
\usepackage{booktabs}%
\usepackage{listings}%
\usepackage{comment}
\usepackage[numbers,sort&compress]{natbib}
\usepackage{algorithmic}
\usepackage{textcomp}
\usepackage{appendix}
\usepackage{subcaption}
\usepackage{hyperref, balance}
\usepackage{dblfloatfix}
\usepackage{pgfplots}
\usepackage{tabularx}
\usepackage{float}     
\usepackage{svg}
\usepackage{makecell}
\usepackage{enumitem}
\usepackage{pifont}
\usepackage{arydshln}
\usepackage{colortbl}
\usepackage{forest}
\usepackage{ragged2e}
\usepackage{wasysym}
\usepackage{diagbox}
\usepackage{forest}
\usepackage[ruled,vlined]{algorithm2e}
\usepackage{setspace}

\theoremstyle{thmstyleone}%
\theoremstyle{thmstyletwo}%

\theoremstyle{thmstylethree}%

\def\BibTeX{{\rm B\kern-.05em{\sc i\kern-.025em b}\kern-.08em
    T\kern-.1667em\lower.7ex\hbox{E}\kern-.125emX}}

\pgfplotsset{compat=1.18}
\begin{document}

\title{Quantum Adversarial Machine Learning: From Classical Adaptations to Quantum-Native Methods $^*$}



\author*[1]{Roozbeh Razavi-Far}\email{roozbeh.razavi-far@unb.ca}
\author[1]{Mohammad Meymani}
\author[2]{Erfan Mahmoudinia}
\author[3]{Dorsa Vazirzade}
\author[4]{Peyman Paknezhad}
\author[5]{Fateme Ghasemi}
\author[5]{Saeed Saravani}
\author[6]{Somayeh Nikkhoo}
\author[7]{Kimia Haghjooei}


\affil[1]{Faculty of Computer Science, University of New Brunswick, Fredericton, Canada}

\affil[2]{Department of Electrical Engineering, Amirkabir University of Technology, Tehran, Iran}

\affil[3]{Faculty of Mathematical and Computer Science, Kharazmi University, Tehran, Iran}

\affil[4]{Pázmány Péter Catholic University, Budapest, Hungary}

\affil[5]{Department of Computer Engineering, Amirkabir University of Technology, Tehran, Iran}

\affil[6]{Department of Computer Engineering, Ferdowsi University of Mashhad, Mashhad, Iran}

\affil[7]{Department of Computer Science, Tarbiat Modares University, Tehran, Iran}


\begingroup
\renewcommand{\thefootnote}{\fnsymbol{footnote}}
\footnotetext[1]{This article has been published in \textit{Artificial Intelligence Review} (Springer Nature). It is distributed under the Creative Commons Attribution 4.0 International License (CC BY 4.0). The final authenticated version is available at: \url{https://doi.org/10.1007/s10462-026-11578-7}}
\endgroup

\abstract{Machine learning has revolutionized numerous industrial domains. Despite recent advances, machine learning models remain vulnerable to adversarial threats. Adversarial machine learning is a field that studies these vulnerabilities to build robust machine learning models. Quantum machine learning is an interdisciplinary field that bridges quantum computing and classical machine learning. While quantum machine learning shows potentials to outperform classical machine learning in complex tasks such as regression, classification, and generative modeling, it remains vulnerable to adversarial attacks. Given the recent advancements in quantum computing and machine learning, the quantum adversarial machine learning field has emerged to study the vulnerabilities of quantum machine learning, possible attacks, and novel quantum-enhanced defense strategies. In this survey, we provide a detailed overview on quantum adversarial machine learning and explore the existing attacks and countermeasures. We also review the theoretical underpinnings of this area, emerging trends, and critical challenges.}




\maketitle

\section{Introduction}
Machine learning (ML) has empowered many areas such as computer vision (CV), finance, healthcare, cybersecurity, and natural language processing (NLP) \cite{shinde2018review}. There has been a growing trend of adversarial threats against ML models, driven by the increasing adoption of these models across various domains. These threats aim to make use of the vulnerabilities that are associated with these models to degrade the performance (i.e., harm their integrity and availability) of the models or to violate their privacy. The study of adversarial threats and designing countermeasures to make the ML models robust against theses threats is called \underline{a}dversarial \underline{m}achine \underline{l}earning (AML) \cite{pitropakis2019taxonomy,kurakin2016adversarial,kaur2022trustworthy,vashagh2026recent,meymani2026divided}. Moreover, these threats have globally impacted the governments and industries. In response, Bipartisan artificial intelligence (AI) task force of USA \cite{AI_Task_Force_2024} and the Department of Homeland Security (DHS) \cite{DHS_AI_Threat_2023} have labeled evasion attacks and generative AI as national threats to individuals' privacy and security, emphasizing the urge for trustworthy AI and robust ML models.

Quantum computing uses the principles from quantum physics, such as superposition and entanglement, {to potentially provide computational advantages over classical methods for certain problem classes} \cite{rietsche2022quantum,sooksatra2021evaluating}. Quantum machine learning (QML) combines quantum computing with machine learning techniques to {potentially offer computational benefits for specific tasks, depending on the encoding strategy, model design, and hardware capabilities.} {Despite promising early results in controlled settings, QML models, similar to classical machine learning (CML)} models, show vulnerabilities towards adversarial threats \cite{ghosh2025adversarial,franco2024predominant,west2023towards, lu2020quantum, Zhang2025Experimental,tian2023recent,xiao2023practical}. For instance, in an evasion attack, the adversary can craft an adversarial sample, which is imperceptible to humans, from the original input that can easily fool the model and cause misclassification; or the adversary may corrupt the training data to degrade the model's performance at test time. Additionally, the adversary might probe the model to discover its learned parameters or to determine whether specific samples were used during the training phase \cite{pitropakis2019taxonomy,kurakin2016adversarial,kaur2022trustworthy,tian2022comprehensive}.

QML models' fundamental vulnerabilities arise from the natural vulnerabilities of CML models to adversarial attacks. Running models on quantum devices has introduced new opportunities for the adversaries to launch attacks compared to CML models. These new vulnerabilities are associated with quantum devices and environments. Attack surface on QML is divided into four main categories: input level, circuit level, measurement level, and hardware level \cite{ghosh2025adversarial,Arias2023Survey, franco2024predominant, Edwards2020Status}. Input level and measurement level are just similar to CML, where the adversary launches attacks such as evasion, exploratory, and poisoning during different stages of model's lifecycle, including training and test/inference time. The circuit embedded in the quantum devices is the first reason that the attack surface of QML models are extended compared to CML models; so the adversary can exploit vulnerabilities of the circuit and attack the logical structure of the model \cite{ghosh2025adversarial,Arias2023Survey, franco2024predominant, Edwards2020Status}. Special hardware components of quantum devices, such as their processors, further increase the attack surface of QML devices, since the adversary leverages vulnerabilities associated with the physical infrastructure, making these attacks particularly challenging to mitigate \cite{Arias2023Survey, franco2024predominant, Xu2023Classification}.

The severity of these threats on QML has motivated the researchers to investigate these vulnerabilities. The study of adversarial attacks and designing countermeasures is known to as \underline{q}uantum \underline{a}dversarial \underline{m}achine \underline{l}earning (QAML). Since QML shares similar characteristics to CML, some attacks and defenses on CML can be transferred to quantum domain. However, classical attacks and defenses that are transferred to QML have shown {varying levels of effectiveness, often depending on the specific model architecture and encoding strategy,} which has motivated the researchers to study and design quantum-native attacks and defenses that make use of the properties of quantum environments \cite{ghosh2025adversarial,west2023towards,yocam2024quantum}.

{To better position QAML within the broader QML ecosystem, recent advances in QML provide important context for QAML. For instance, hybrid QML models have shown task-specific performance gains in controlled settings, while reinforcement learning-based approaches highlight potential advantages in quantum sensing under idealized assumptions. In parallel, variational and generative QML frameworks demonstrate improved expressivity and parameter efficiency, but also expose challenges related to scalability and training stability \cite{xiao2023practical,xu2025toward,xiao2023quantum}. These results suggest that QML advantages are highly context-dependent, highlighting the need to study robustness within realistic settings.}

{Table \ref{tab:compare2other-surveys} provides a comprehensive comparison between our survey and existing surveys in the literature, clearly highlighting the distinctive contributions and novel aspects of our work.}
\begin{table*}[htbp]
    \centering
    \caption{Comparative analysis of existing surveys and our work - evasion (Eva), data poisoning (DP), oracle-based (Ora), hardware- and circuit-level (HCL), application specific (App), classical adaptation (CA), and Quantum Specific (QS).}
    \resizebox{\linewidth}{!}{    \begin{tabular}{l | c | c c c c c | c c c | c | c | c | c}
    \hline
    Survey &\makecell{Threat\\Model}  & \multicolumn{5}{c|}{Attacks} & \multicolumn{3}{c|}{Defenses} & \makecell{Theoretical\\Perspectives} & Datasets & \makecell{Evaluation\\Metrics} & \makecell{Simulation\\Frameworks}\\
    &&\rotatebox{270}{Eva}&\rotatebox{270}{DP}&\rotatebox{270}{Ora}&\rotatebox{270}{HCL}&\rotatebox{270}{App}&\rotatebox{270}{CA}&\rotatebox{270}{QS}&\rotatebox{270}{HCL}&&&&\\
    \hline
    [1]&\color{blue}\checkmark& &\color{blue}\checkmark&\color{blue}\checkmark&\color{blue}\checkmark& & &\color{blue}\checkmark&\color{blue}\checkmark& & & & \\
    
    [2]& & & & &\color{blue}\checkmark& &\color{blue}\checkmark& & & & & & \\
    
    [3]& &\color{blue}\checkmark&\color{blue}\checkmark&\color{blue}\checkmark& & & &\color{blue}\checkmark& & &\color{blue}\checkmark& & \\
    
    [4]& &\color{blue}\checkmark& &\color{blue}\checkmark& & &\color{blue}\checkmark& & & & & & \\
    
    This survey   & \color{blue}\checkmark & \color{blue}\checkmark & \color{blue}\checkmark & \color{blue}\checkmark & \color{blue}\checkmark & \color{blue}\checkmark & \color{blue}\checkmark& \color{blue}\checkmark & \color{blue}\checkmark & \color{blue}\checkmark & \color{blue}\checkmark& \color{blue}\checkmark & \color{blue}\checkmark\\
    \hline
    \end{tabular}}
\label{tab:compare2other-surveys}
\end{table*}

In this survey, we bridge the gap between classical AML and QAML, underscoring the types, limitations, challenges, and the future of QAML. Our contributions are as follows:
\begin{itemize}
    \item We introduce a novel threat model for QAML, which is derived from the classical threat model on AML.
    \item We introduce and categorize the adversarial threats on QML, distinguishing those attacks transferred from CML and the quantum-native ones that are emerged due to the increased attack surface of quantum systems.
    \item We introduce the existing countermeasures against the adversarial threats, highlighting their effectiveness and limitations.
    \item We examine the theoretical principals of QAML, analyzing the behavior of the quantum models in adversarial settings.
    \item We investigate the benchmarks in QAML, underscoring the pros and cons associated with each benchmark category.
\end{itemize}

The rest of the paper is organized as follows. In Section \ref{sec:background}, we give a brief background of QML and AML. In Section \ref{sec:taxonomy-of-threats}, we explain various aspects of the post quantum threat model. In Section \ref{sec:attacks-on-qml}, we investigate different adversarial attacks on quantum ML models. In Section \ref{sec:defenses-in-qml}, we explore the common defense strategies in quantum ML models. In Section \ref{sec:theoretical-perspectives}, we illustrate the theoretical foundations of QML. In Section \ref{sec:benchmark-datasets-evaluation}, we introduce the benchmarks and evaluation metrics in quantum settings. In Section \ref{sec:emerging-trends-and-applications}, we discuss the emerging trends in QAML. In Section \ref{sec:challenges-and-open-problems}, we explore the current challenges and open problems. Finally, in Section \ref{sec:conclusion}, we conclude our work.
\section{\textbf{Background}}
\label{sec:background}
In this section, we provide a brief background on quantum computing basics and different types of QML models.
\subsection{\textbf{Basics of Quantum Computing}}
Quantum computing exploits quantum-mechanical effects to encode and process the information more efficiently than classical computing systems \cite{marella2020introduction}. Its foundation rests on several core principles such as qubits, superposition, entanglement, quantum gates, circuits, unitary operations, decoherence, and quantum noise.

\subsubsection{\textbf{Qubits and Superposition}}
A \underline{qubit} (quantum bit) is the core information carrier in quantum computing systems. Unlike a classical bit, which occupies only one of two distinct states, either 0 or 1, a qubit might exist in a quantum \underline{superposition} of both states simultaneously. This quantum property enables a system of qubits to represent and process multiple computational paths in parallel \cite{marella2020introduction}. For instance, a three-qubit system in superposition can simultaneously encode all $2^3=8$ possible classical states.  Mathematically, a qubit is written as:
    \begin{equation}
    |\rho\rangle = \alpha |0\rangle + \beta |1\rangle 
    \end{equation}
    where $|.\rangle$ denotes a column vector, $\alpha$ and $\beta$ are complex amplitudes with $|\alpha|^2 + |\beta|^2 = 1$, determining the contribution of each vector \cite{chae2024elementary}.

\subsubsection{\textbf{Entanglement}}
    When qubits are entangled, the state of one instantly correlates with another, even across vast distances. This non-classical link means one cannot fully describe each qubit independently. For instance, measuring one will immediately affect its partners' outcomes. Entanglement is a key factor enabling quantum algorithms to secure communication \cite{khan2024quantum}.

\subsubsection{\textbf{Quantum Gates and Circuits}}
Quantum gates manipulate qubits via reversible, unitary operations. Common single-qubit gates include Pauli-X (bit flip), Pauli-Z (phase flip), and the Hadamard gate (creating superposition) \cite{marella2020introduction}. Two-qubit gates, like the controlled NOT (CNOT), generate entanglement \cite{hughes2021quantum}. Quantum circuits sequence these gates on qubit registers to perform computation. Unlike the classical logic (e.g., AND and OR), quantum gates must obey unitarity, ensuring the computation is reversible and the total probability distribution is preserved \cite{munoz2022everything}.

\subsubsection{\textbf{Unitary Evolution}}
Any transformation on qubits can be represented by a unitary matrix $U$ satisfying $U^\dagger U=I$, ensuring the operation is reversible and probability distribution is preserved. Designing algorithms involves chaining these unitaries to guide the system toward a solution \cite{palao2002quantum}.

\subsubsection{\textbf{Wavefunction Collapse}}
    At the end of the computation, qubits must be measured. This action collapses the superposition into a definite state, yielding 0 or 1 with probabilities $|\alpha|^2$ or $|\beta|^2$. Importantly, measuring destroys superposition, collapsing the system irreversibly \cite{chae2024elementary}.

\subsubsection{\textbf{Decoherence and Quantum Noise}}
    Decoherence occurs when qubits interact with their environment, through thermal fluctuations, electromagnetic noise, or cosmic rays, collapsing delicate superposition and entanglement \cite{schlosshauer2019quantum}.
    Typical error rates in qubits range from $10^{-2}$ to $10^{-4}$, vastly higher than error rates in classical systems ($\approx10^{-13}~\text{to}~10^{-18}$) \cite{chae2024elementary}.
    To mitigate decoherence, current quantum computing platforms, particularly those with superconducting qubits, operate at temperatures near to absolute zero to suppress thermal noise and maintain coherence \cite{kjaergaard2020superconducting}.

\subsection{\textbf{Overview of Quantum Machine Learning}}
Quantum ML models, similar to CML, can be generally categorized into supervised, unsupervised, and reinforcement learning paradigms. These models use quantum transformations and measurements to assign labels, extract patterns from the encoded data, or learn by interacting with a quantum environment.
\subsubsection{\textbf{Supervised QML}}
It involves learning a mapping from input to output using labeled data. Prominent examples include Quantum Support Vector Machines (QSVMs) and Quantum Neural Networks (QNNs).

\textbf{QSVMs} \cite{suzuki2024quantum} employ quantum kernels to efficiently compute inner products between encoded data points in a high-dimensional Hilbert space. Experimental implementations using trapped-ion processors {have shown competitive performance in certain experimental settings, although these results are typically limited to small-scale problems and specific hardware configurations.}

\textbf{QNNs} \cite{jeswal2019recent} are built on parameterized quantum circuits designed to model complex relationships between input features by transforming them into high-dimensional quantum states. These transformations enable the circuit to act as a powerful learning model, using the expressivity of quantum operations to encode nonlinear decision boundaries. QNNs typically use hybrid optimization schemes, where classical algorithms adjust quantum gate parameters based on measurements from quantum hardware \cite{schuld2019quantum,jeswal2019recent,upadhyay2025quantum}. However, their scalability is hindered by the barren plateau problem, in which gradients of the cost function diminish exponentially as the circuit depth or number of qubits increases. This phenomenon makes optimization increasingly difficult, especially in randomly initialized or highly entangled circuits, posing a fundamental challenge to training deep quantum models \cite{mcclean2018barren}.

\subsubsection{\textbf{Unsupervised QML}}
It focuses on discovering patterns in unlabeled data. A representative unsupervised algorithm is quantum k-means clustering.

\textbf{Quantum k‑means} is a quantum-enhanced clustering algorithm that adapts the classical k-means process to quantum computing frameworks. It uses quantum parallelism to speed up key steps such as distance calculations and cluster assignment. Data is first encoded into quantum states, allowing a quantum circuit to compute similarities between points and centroids simultaneously. Using quantum subroutines like the SWAP test or Grover’s search, each data point is assigned to the nearest cluster. The centroids are then updated with the help of classical post-processing, and the process repeats until the clusters stabilize. Quantum k-means {has the potential to reduce runtime under certain assumptions, although practical advantages remain limited in current Noisy Intermediate Scale Quantum (NISQ) implementations} \cite{kerenidis2019q,poggiali2024quantum}.
\subsubsection{\textbf{Quantum Reinforcement Learning (QRL)}}
It is an emerging area, where quantum agents learn to make sequential decisions through interactions with an environment. \underline{Quantum deep Q-learning}, for instance, integrates parameterized quantum circuits with classical feedback mechanisms to approximate value functions or policies. Although QRL is in its infant stage of development, these models are promising for tasks involving large or uncertain action spaces due to {potential advantages in exploration, although such benefits remain largely theoretical and require further empirical validation} \cite{skolik2022quantum,meyer2022survey,li2025robust,xu2025toward,xiao2023quantum}.

\underline{V}ariational \underline{q}uantum \underline{c}ircuits (VQCs), also known as \underline{p}arameterized \underline{q}uantum \underline{c}ircuits (PQCs), are central to many QML models and include tunable parameters that can be optimized during training. These circuits encode classical inputs and iteratively evolve their parameters to minimize a loss function, guided by classical optimizers. VQCs form the computational backbone of models like QSVMs, QNNs, and other discriminative or generative quantum classifiers \cite{li2022recent}. Their adaptability and noise robustness make them especially suited for near-term quantum devices, and they have demonstrated competitive performance in various applications including image classification and generative modeling \cite{lu2020quantum,xiao2023quantum}. 

\textbf{Hybrid quantum-classical architectures} are broadly employed in quantum machine learning, where PQCs perform transformations on quantum data, while classical optimizers guide the training process. This method, which is central to \underline{v}ariational \underline{q}uantum \underline{a}lgorithms (VQAs), allows near-term quantum hardware to handle useful tasks despite current limitations. The classical optimizer updates quantum parameters based on cost function outputs from the quantum circuit, forming an iterative feedback loop. These hybrid systems strike a practical balance between quantum expressivity and classical computational control \cite{li2022recent,li2022quantum,huang2023image}.
\subsection{\textbf{QAML Variants}}
While QAML studies the vulnerabilities associated with the QML models and explores adversarial attacks and defenses for these models, there are closely related quantum adversarial domains, where the goal is to defend the classical models against quantum-powered or quantum-inspired adversaries. These areas are known as post-quantum adversarial machine learning (PQAML) and quantum-inspired adversarial machine learning (QiAML).

PQAML is an emerging field that considers the adversaries equipped with quantum computing capabilities. This field aims to secure the CML models against the threats posed by quantum-powered adversaries in post-quantum era. PQAML differs from QAML, which instead focuses on the threats and defenses of QML models themselves. Post-quantum cryptography plays an important role by providing quantum-resistant cryptographic foundations that can be combined with secure ML pipelines to ensure long-term protection of data and model's parameters against quantum adversaries \cite{tan2023post,darzi2024pqc}. 

Quantum-inspired adversarial machine learning (QiAML), on the other hand, aims to develop CML models by incorporating quantum principles into CML algorithms. QiAML studies QiML models to build attacks and defenses that make use of quantum principles. The integration of quantum principles and CML models has {enabled the development of potentially more expressive attack and defense strategies, though their effectiveness varies depending on the application and experimental setup.} Additionally, those defenses that are built on quantum-inspired algorithms have shown stronger resilience against adversarial threats \cite{huynh2023quantum,kejriwal2025advancing,tseng2024ai}. Table \ref{tab:qml-variants} illustrates differnt aspects of each variant.
\begin{table}
    \centering
    \caption{Comparison between AML, QAML, PQAML, and QiAML - quantum (Q) and classical (C) models.}
    \begin{tabular}{l c c c c c}
        \hline
        Variant & \multicolumn{2}{c}{\makecell{Target Model}} & \multicolumn{2}{c}{\makecell{Threat Model}}&Birth Year\\
        &C&Q&C&Q&\\
        \hline
        AML & \color{green}\checkmark & \color{red}\ding{55} & \color{green}\checkmark&\color{red}\ding{55}&2004\\
        
        QAML & \color{red}\ding{55} & \color{green}\checkmark & \color{green}\checkmark&\color{green}\checkmark&2019\\
        
        PQAML & \color{green}\checkmark & \color{red}\ding{55} & \color{red}\ding{55}&\color{green}\checkmark&2024\\
        
        QiAML & \color{green}\checkmark & \color{red}\ding{55} & \color{green}\checkmark&\color{red}\ding{55}&2023\\
        \hline
    \end{tabular}
    \label{tab:qml-variants}
\end{table}

In this survey, we merely focus on QAML defenses and attacks since PQAML and QiAML domains are still emerging and research works in these areas are scarce.
\section{\textbf{Taxonomy of Quantum Adversarial Threats}}
\label{sec:taxonomy-of-threats}
\begin{figure}[htbp]
    \centering
    \includegraphics[width=0.8\linewidth]{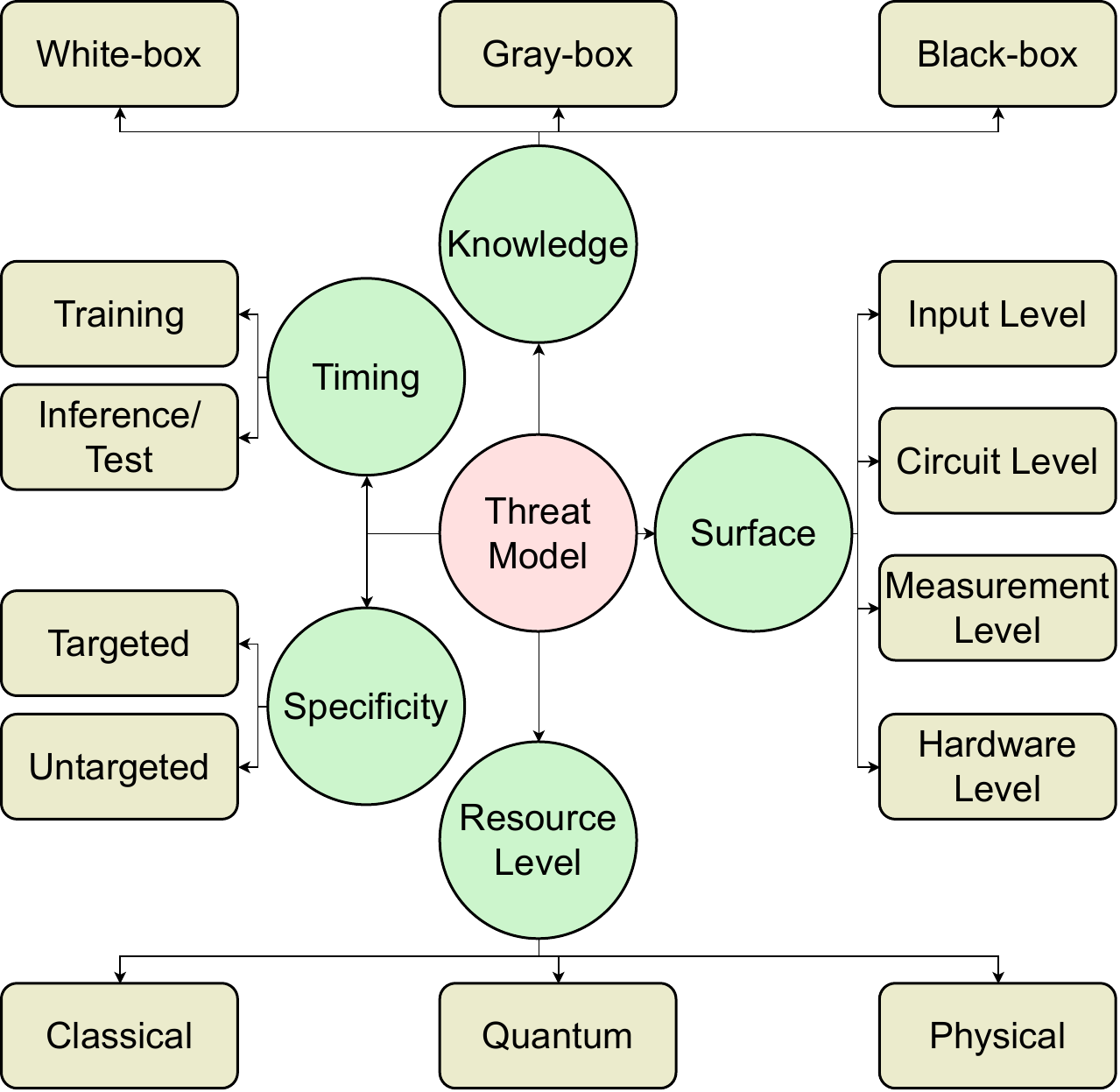}
    \vspace{1em}
    \caption{Quantum adversarial threat Model.}
    \label{fig:threat-model}
\end{figure}

In this section, we introduce the quantum adversarial threat model, categorizing adversary types from different angels. Figure \ref{fig:threat-model} shows the threat model in a quantum setting.

As QML systems advance toward real-world deployment, understanding their security and privacy vulnerabilities becomes a critical challenge \cite{marchiori2025attaq,west2023towards,franco2024predominant,Edwards2020Status}. QML models introduce a uniquely expanded threat landscape, combining vulnerabilities from classical software frameworks with novel attack vectors that target the quantum computations and the physical hardware itself \cite{Arias2023Survey,franco2024predominant,Xu2023Classification}. To systematically navigate this complex domain, a structured taxonomy is essential for categorizing and analyzing these diverse threats. This taxonomy is organized along five fundamental dimensions: \underline{timing}, \underline{knowledge}, \underline{surface}, \underline{resource level}, and \underline{specificity}.

\subsection{\textbf{Attack Surface}}
This dimension specifies where in the QML pipeline the adversary launches its attacks. QML systems present a uniquely broad attack surface, from classical inputs to quantum hardware itself \cite{Arias2023Survey, franco2024predominant, Edwards2020Status}. There are four primary surfaces:

\textbf{Input-level surface} is the most common category, which involves adding crafted perturbations to the classical data or the encoded quantum state to cause misclassification at test/inference time or poison the training set to degrade the general performance of the model \cite{lu2020quantum, west2023towards, liu2020vulnerability, Li2025Computable, Georgiou2025Generalization, Ren2022Experimental, montalbano2025quantum, kundu2025adversarial}.

\textbf{Circuit-level surface} targets the logical structure of the model, where an adversary can exploit compiler vulnerabilities \cite{Arias2023Survey}, embed malicious logic like a quantum Trojan into the circuit's architecture \cite{franco2024predominant}, or attempt to reverse-engineer the protected intellectual property \cite{das2023randomized}.

\textbf{Measurement-level surface} exploits the final stage of computation, where an attacker can use readout errors for information leakage or computation reliability \cite{franco2024predominant,Shi2024QuanTest}.   

\textbf{Hardware-level surface} is the most physically tangible surface, which involves targeting the quantum processor and its control infrastructure by inducing hardware noise and errors and exploiting inherent physical weaknesses \cite{franco2024predominant, Arias2023Survey}.
\subsection{\textbf{Attacker's Capabilities and Knowledge}}
This dimension defines the power and knowledge of the attacker that are crucial for establishing a realistic threat model \cite{liu2021privacy, Edwards2020Status}. We categorize these capabilities along two main axes: the attacker's level of access (knowledge) to the model and the computational resources they possess.
\subsubsection{\textbf{Knowledge}}
This describes the attacker's degree of knowledge about the target model's internal mechanisms including design, operations, and parameters.

\textbf{White-box}: The attacker has perfect knowledge of the model, including its architecture, parameters, loss function, gradients, and training data \cite{liu2021privacy,west2023towards,lu2020quantum,vashagh2026recent}. This powerful assumption is often used to probe a model's fundamental vulnerabilities and is a prerequisite for many gradient-based attacks like projected gradient descent (PGD) and fast gradient sign method (FGSM) \cite{goodfellow2014explaining, madry2017towards,yocam2024quantum, montalbano2025quantum, ElMaouaki2024RobQuNNs}.

\textbf{Black-box:} The attacker has no internal knowledge and can only query the model to observe its input-output behavior \cite{liu2021privacy, west2023towards, lu2020quantum,meymani2026divided}. This is a more practical scenario, where the primary technique is the \textit{transfer attack}. In this attack, the adversarial examples are generated against a surrogate model, and, then, used against a target quantum model to degrade its inference-time performance \cite{lu2020quantum, west2023benchmarking, ElMaouaki2024RobQuNNs,papadopoulos2025numerical}. Gradient-free methods, such as genetic algorithms, are also employed for black-box attacks on real hardware \cite{Jin2025Realizing}.

\textbf{Gray-box:} This represents a practical intermediate scenario, where an attacker has partial knowledge. For instance, a malicious insider at a quantum cloud provider might know the victim's data encoding circuit but has no information about the specific architecture of their parameterized quantum circuit \cite{kundu2025adversarial,tseng2024ai,meymani2026divided}.
\subsubsection{\textbf{Resource Level}}
This distinguishes adversaries based on their computational power. This is a critical dimension in QAML as it defines the boundary between current and future threats.

\textbf{Classical Adversary:} The attacker is limited to classical computing resources to analyze the model and craft attacks \cite{Arias2023Survey, west2023towards}. The vast majority of current QAML research assumes a classical adversary.   

\textbf{Quantum Adversary:} It's a more powerful threat model, where the adversary has its own quantum computer \cite{ west2023towards}. This would allow them to craft sophisticated quantum states for attacks or potentially break the classical cryptography that protects a hybrid quantum-classical architecture.    

\textbf{Physical Adversary:} This capability assumes the attacker can physically interact with the quantum computer to induce faults, for instance by using Electromagnetic (EM) radiation or thermal manipulation, the adversary can craft a distinct and potent threat vector \cite{Arias2023Survey, Xu2023Classification}.
\subsection{\textbf{Attack Timing}}
Adversarial attacks can be categorized based on the time they occur w.r.t the machine learning lifecycle \cite{liu2021privacy}. This dimension is divided into two primary phases: the training phase, where the model is fitted to the data, and the inference phase, where the trained model makes predictions.

\textbf{Training-time:} Attacks during the training phase include various types such as poisoning, backdoor, fault injection, among others, which compromise the model during its training session \cite{liu2021privacy, Wiebe2017Hardening, Onim2025Detection}. The adversary's objective is to corrupt the final learned model by manipulating the training data \cite{kundu2025adversarial, Onim2025Detection} or the model's logic itself \cite{franco2024predominant, Wiebe2017Hardening}. This can be done to degrade the model's overall accuracy or to embed a hidden backdoor, such as a Quantum Trojan, that can be activated later \cite{franco2024predominant, liu2021privacy,das2023trojannet}. The first systematic data poisoning attack specifically designed for QML, named quantum indiscriminate data poisoning attack (QUID), demonstrated the viability of this threat in the NISQ era \cite{kundu2025adversarial}.

\textbf{Test/Inference-time:} Attacks during this phase mainly include evasion and oracle-based attacks. These occur after a model has been fully trained and deployed \cite{lu2020quantum, yocam2024quantum, liu2021privacy}. Here, the adversary's goal is to craft a malicious input or an adversarial example, that causes the operational model to make an incorrect prediction \cite{west2023towards, montalbano2025quantum, Ren2022Experimental}. This is the most widely studied attack scenario in QAML, forming the basis of nearly all experimental works on adversarial robustness \cite{lu2020quantum, Ren2022Experimental, Zhang2025Experimental, Jin2025Realizing}. Privacy attacks, such as model extraction and membership inference, also target the deployed model in order to steal the functionality and/or private properties of the model \cite{liu2021privacy, Ghosh2024Reverse,kundu2024evaluating,upadhyay2025quantum}.   

\subsection{\textbf{Specificity}}
Based on specificity, adversarial attacks could be categorized as targeted or untargeted (indiscriminate).

\textbf{Untargeted Attacks:} The adversary's objective is to cause any type of error, simply to make the model fail \cite{lu2020quantum, west2023towards, montalbano2025quantum, akter2023exploring}. This is typically achieved by crafting a perturbation that maximizes the model's loss function, thereby pushing the input across any nearby decision boundary without any concern for the final incorrect output \cite{west2023benchmarking, Wu2023Radio, Wendlinger2024Comparative}.

\textbf{Targeted Attacks:} These are more sophisticated, aiming to deceive the model into producing a specific, predetermined incorrect output \cite{lu2020quantum, yocam2024quantum}. A classic example is manipulating an image of a handwritten \reflectbox{'}7' to be exclusively classified as a \reflectbox{'}9' \cite{lu2020quantum}. A Quantum Trojan backdoor exemplifies a powerful targeted attack, as it is designed to provide the attacker with high-success-rate control over the model's output when activated \cite{franco2024predominant}.   

\subsection{Operationalization of the QAML Threat Taxonomy}
{While the proposed taxonomy provides a structured classification of adversarial threats in QAML, it can also be used as a practical guideline for experimental design. In particular, researchers can define threat models by specifying key dimensions such as attack type, attacker's knowledge, access level, perturbation constraint, and attack objectives. These definitions can then guide the selection of experimental configurations, including datasets, encoding strategies, and circuit parameters, as reflected in Tables \ref{tab:attack-research-works} and \ref{tab:defense-research-works}. A typical evaluation workflow consists of defining the threat model, implementing the corresponding attack, and assessing the model's robustness under the specified conditions while reporting sufficient experimental and architectural details to ensure reproducibility. This connection between taxonomy and evaluation provides a lightweight yet practical framework for conducting and comparing QAML studies.

Building on this structured threat modeling process, a complete QAML evaluation should include several complementary dimensions to ensure meaningful and reproducible results. In particular, robustness should be assessed using metrics such as attack success rate under clearly defined perturbation constraints and a clear threat model, together with the trade-off between clean accuracy and robustness. For black-box scenarios, query complexity should also be reported to identify the efficiency and practicality of attacks. 

Furthermore, quantum-specific factors must be explicitly considered, including the choice of noise models and whether they are synthetic or calibrated to real hardware. It is also necessary to distinguish between results obtained from simulation frameworks and those validated on real quantum devices, as performance may differ notably across these environments. To ensure reproducibility and fairness, experimental specifications need to be reported in sufficient detail, including the encoding strategy, the number of qubits, the circuit depth, and the optimization settings. All together, these components form a practical evaluation framework that complements the proposed taxonomy and supports more standardized QAML experiments.}
\section{\textbf{Adversarial Attacks on Quantum Models}}
\label{sec:attacks-on-qml}
\begin{figure}[htbp]
    \centering
    \includegraphics[width=0.8\linewidth]{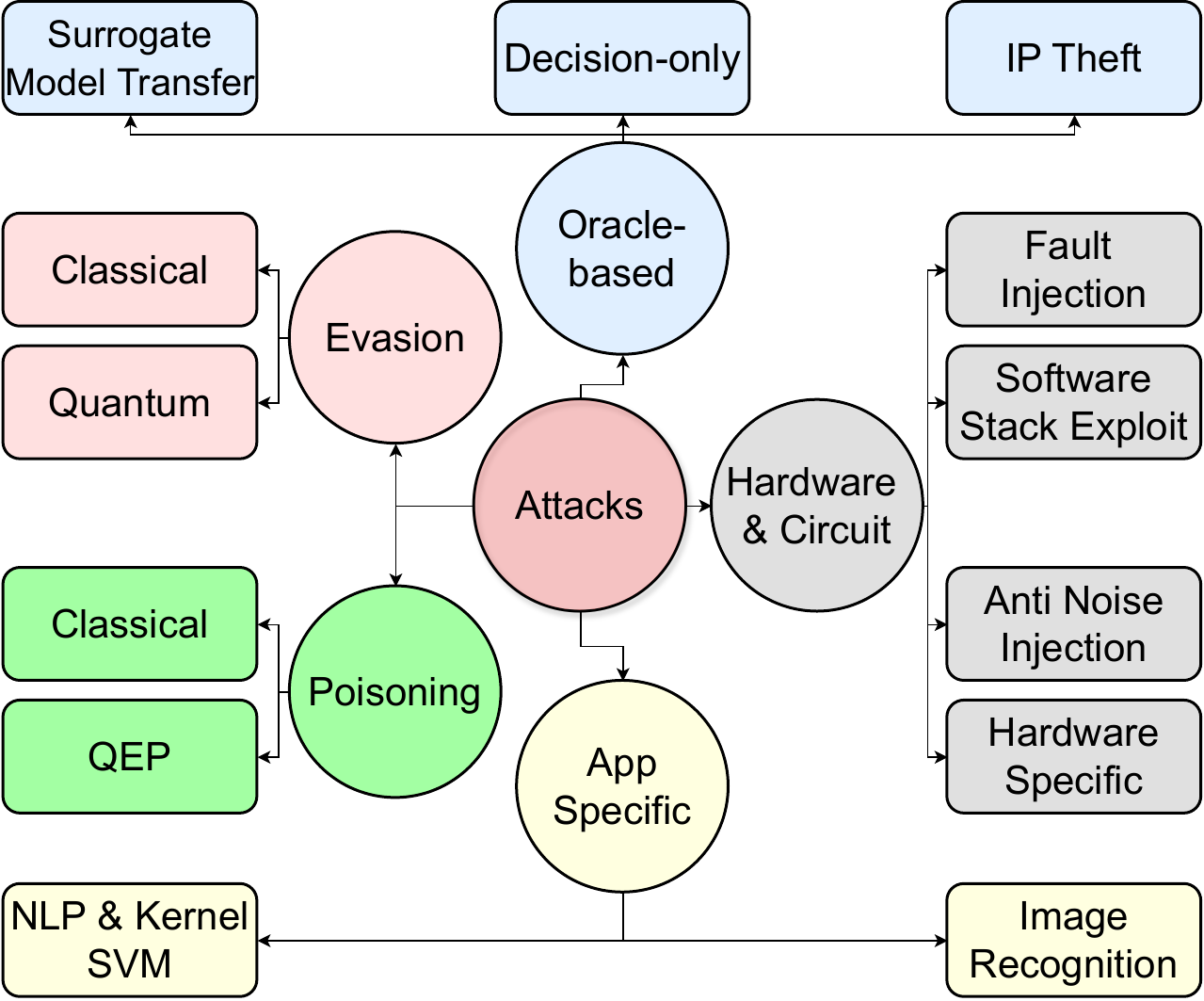}
    \caption{General taxonomy of adversarial attacks in QML - quantum encoding poisoning (QEP), surrogate model transfer (SMT), and software stack exploitation (SSE).}
    \label{fig:attack-types}
\end{figure}

In this section, we introduce the attacks transferred from CML to QML, as well as, the new attacks stemmed from the quantum environment. The general taxonomy of theses attacks are shown in Figure \ref{fig:attack-types}.
\subsection{\textbf{Evasion}}
Evasion attacks on QML represent the most extensively studied category of threats in the QAML literature. The adversary's fundamental goal is to deceive a QML model by adding a small, often imperceptible perturbations to legitimate input samples, thereby causing a misclassification \cite{west2023towards, lu2020quantum, Zhang2025Experimental,baral2025adversarial}. These attacks typically occur during test/inference, where the objective is to fool a trained model. The core mechanism involves crafting a perturbation, $\delta$, that is just large enough to push the input sample across a learned decision boundary in the high-dimensional feature space \cite{montalbano2025quantum, lu2020quantum,yocam2024quantum}.
\subsubsection{\textbf{Classical Perturbation}}
The noise is added to the classical data vector, $x$, before it is encoded into the quantum state. The resulting adversarial example $x_{adv}$ is given by the following relation:
\begin{equation}
x_{adv} = x + \delta   
\end{equation}

The perturbation $\delta$ is constrained to be small by bounding its magnitude under a suitable $l_p$-norm, such that $||\delta||_p \le \epsilon$, where $\epsilon$ is the perturbation bound \cite{Li2025Computable, Georgiou2025Generalization}.
A variety of algorithms, many adapted from the classical domain, are used to find an effective perturbation $\delta$. The most common white-box, gradient-based algorithm is FGSM \cite{yocam2024quantum, Tiwo2025Financial, ElMaouaki2024AdvQuNN,wang2023defending}. For a loss function $\ell(\theta, x, y)$, FGSM computes the perturbation in a single step \cite{lu2020quantum}:
\begin{equation}
x_{adv } = x + \epsilon \cdot \mathrm{sign}(\nabla_x \ell(\theta, x, y))  
\end{equation}
where $\epsilon$ is the perturbation size, $\theta$ denotes the parameters of the model, and $\mathrm{sign}(.)$ returns the sign of an input. More powerful, iterative versions like PGD or the Basic Iterative Method (BIM) \cite{kurakin2018adversarial} perform this process over multiple smaller steps to find a more optimal adversarial example \cite{lu2020quantum, Wendlinger2024Comparative, Winderl2023Depolarization}. To better handle the geometry of quantum state space, quantum-native attacks like trace distance-PGD (TD-PGD) have also been developed \cite{Li2025Computable}. Beyond gradient-based methods, black-box attacks using techniques like genetic algorithms have been successfully demonstrated on real quantum hardware \cite{Jin2025Realizing}.

The threat of evasion attacks has been {demonstrated across a wide range of models and experimental settings}. These include variational classifiers for medical images and handwritten digits \cite{Ren2022Experimental}, quantum kernel methods \cite{montalbano2025quantum}, quanvolutional neural networks \cite{ElMaouaki2024RobQuNNs}, classifiers for radio signals \cite{Wu2023Radio}, and quantum phases of matter \cite{lu2020quantum}. More advanced threats like \underline{universal adversarial perturbations}, where a single noisy pattern can corrupt various inputs, have also been proven to exist for quantum classifiers \cite{gong2022universal,qiu2023universal}, highlighting the severity of this vulnerability.

\subsubsection{\textbf{Quantum Perturbation}}
The noise is directly applied to the quantum state, $\rho(x)$, after the data has been encoded. This represents a quantum-native attack, where the perturbed state, $\rho'$, must remain close to the original state. This closeness is bounded by a quantum-specific distance metric, such as the trace distance, $T(\rho, \rho') \le \epsilon$ \cite{Georgiou2024Information,du2021quantum}, or by ensuring a high fidelity, $F(\rho, \rho') \ge 1-\epsilon$ \cite{Li2025Computable, weber2021optimal}.

In white‐box scenario, an adversary has full access to the quantum model’s inner mechanisms, including the model's state‐preparation circuits, variational parameters and loss function, and uses gradient information together with the geometry of quantum state space to craft the smallest possible disturbance that still flips the classifier’s decision. By examining one‐step methods, iterative schemes and constrained‐optimization techniques, this category uncovers fundamental weaknesses in QML representations and sets a benchmark for defense evaluations. In practice, these attacks measure how slight a perturbation must be to guarantee misclassification, guiding the design of more robust encoding schemes and training procedures with certified guarantees.

Lu \emph{et al.} \cite{lu2020quantum} adapted the classical FGSM to quantum models by using the parameter‐shift rule for a single‐qubit rotation \(R(\theta)=e^{\displaystyle -i\cdot\theta\cdot P/2}\),
\begin{equation}
\frac{\partial \ell}{\partial \theta}
= \frac{\ell\bigl(\theta+\tfrac{\pi}{2}\bigr)\;-\;\ell\bigl(\theta-\tfrac{\pi}{2}\bigr)}{2}\,,   
\end{equation}
to estimate the gradient \(\nabla_{\rho}\ell\), where $P$ is the Pauli operator. They then apply,
\begin{equation}
\rho' =
e^{\displaystyle-i\cdot\epsilon\cdot\mathrm{sign}(\nabla_{\rho}\ell)\cdot H}\times\rho\times
e^{\displaystyle i\cdot\epsilon\cdot\mathrm{sign}(\nabla_{\rho}\ell)\cdot H},    
\label{eq:quantum-FGSM}
\end{equation}
where \(H\) is an appropriate Hermitian generator \cite{lu2020quantum,west2023towards}.  FGSM’s main advantage is that it only needs two circuit evaluations per gradient component. However, since FGSM is a single‐step attack, its effectiveness can drop under realistic noise and normalization constraints. 

To increase attack's success, PGD over density operators has been proposed \cite{montalbano2025quantum,gong2024randomized}.  At each iteration \(k\):
\begin{align}
&\rho_{k+1} = \Pi_{\mathcal{B}}\bigl(\rho_k -\alpha\cdot\nabla_{\rho}\ell(\rho_k)\bigr),\\
&\mathcal{B} = \{\sigma : D_{\mathrm{tr}}(\rho_0,\sigma)\le\varepsilon\},
\end{align}
where $\alpha$ is the step size, $\sigma$ is any possible quantum state inside the feasible set $\mathcal{B}$, $D_{\mathrm{tr}}$ is the trace-distance, and \(\Pi_{\mathcal{B}}\) projects the perturbed sample back onto the trace‐distance ball of radius \(\epsilon\). Although PGD can achieve high success rates in ideal simulators, its query cost scales as \(O(K \cdot m)\), where \(m\) is the number of circuit evaluations per iteration and $K$ is the total number of iterations. The effectiveness of $m$ degrades under realistic noise.

CW‐style attacks frame the problem as:
\begin{equation}
\min_{\delta}\;c\,\ell(\rho+\delta)\;+\;D_{\mathrm{tr}}(\rho,\rho+\delta),    
\end{equation}
and solve it via gradient descent plus line search, often with a binary search over the regularization constant \(c\) \cite{yocam2024quantum,kundu2025adversarial}.  By taking into account the curvature of the Bloch sphere, these methods find perturbations of smaller-norm than PGD. However, they demand many more optimization steps.  Typically, \(c\) is searched between \(10^{-3}\) and \(10^{3}\) to balance perturbation size and computational cost. CW finds smaller‑norm perturbations but typically requires substantially more optimization steps compared to PGD \cite{west2023towards}.

{Evasion attacks in QAML are largely derived from classical adversarial machine learning and primarily focus on perturbing encoded data to induce misclassification during inference time. While these methods are effective at demonstrating that QML models inherit vulnerabilities from classical counterparts, their applicability is often tied to assumptions such as access to gradients or model parameters. Moreover, their effectiveness can depend strongly on the encoding strategy and the specifications of the quantum circuit, which introduces variability that is not present in classical settings. As a result, although these attacks establish an important baseline for vulnerability analysis, their behavior across different QML configurations is not yet fully understood.}
\subsection{\textbf{Data Poisoning}}
Poisoning attacks on QML are among the most challenging attacks. Poisoning can occur in both raw training data and quantum encoding. In classical data poisoning attacks, the samples are first poisoned, and, then, are encoded to their quantum state. In quantum encoding poisoning (QEP), unlike classical data poisoning, the adversary poisons the training data after it is encoded into a quantum state. Poisoning attacks on QML, like CML, can use different strategies such as label flipping, backdoors, and so on \cite{kundu2025adversarial,franco2024secqml,guo2025backdoor}. For instance, QUID \cite{kundu2025adversarial} flips the labels of a subset of raw training data by analyzing their corresponding quantum encoding. In \cite{zhao2025black}, the authors propose a black-box backdoor attack that poisons only a small portion of the raw training data by embedding a learned universal adversarial perturbation; the backdoor is triggered during inference, when it is exposed to the same universal adversarial perturbation, which causes a targeted misclassification.

{Data poisoning attacks are less extensively studied compared to evasion attacks, with existing works typically focusing on specific datasets and model configurations. While these approaches demonstrate that QML models are vulnerable during the training phase, the current literature remains limited in providing a unified understanding of their impact across different QML architectures and learning scenarios.}
\subsection{\textbf{Oracle‐Based Attacks}}
When QML models are offered as cloud or edge services, attackers cannot see internal parameters or gradients, and must rely on classical outputs, either the predicted label or class probabilities. Oracle‐based attacks study how much an adversary can achieve with this limited feedback. This line of work is vital for real‐world threats, since it models scenarios where QML is a service and probes the minimal information needed to fool quantum classifiers. Moreover, these attack types align the most with the exploratory attacks in the classical AML settings.
\subsubsection{\textbf{Surrogate‐model transfer}}
These attacks proceed in two steps.  First, the attacker queries the QML API on inputs \(\{\rho_i\}_{i=1}^M\) to collect labels \(y_i = f_{\mathrm{QML}}(\rho_i)\).  Next, they train a classical surrogate model \(f_{\mathrm{surr}}\) by minimizing the following surrogate loss:
\begin{equation}
\mathcal{L}_{\mathrm{surr}}
= \frac{1}{M}\sum_{i=1}^M \ell\bigl(f_{\mathrm{surr}}(\rho_i),\,y_i\bigr), 
\end{equation}
where \(\ell\) stands for the classification loss.  In the second phase, classical PGD crafts perturbations \(\delta\) on surrogate inputs:
\begin{equation}
\rho' = \Pi_{\mathcal{B}}\!\bigl(\rho + \alpha\cdot\mathrm{sign}(\nabla_{\rho}\ell(f_{\mathrm{surr}}(\rho),y))\bigr),   
\end{equation}
and sends \(\rho'\) back to the quantum model. Fewer queries can decrease surrogate accuracy, but save the cost that can boost transfer success until convergence \cite{Wendlinger2024Comparative,lu2020quantum,papernot2017practical}.    
\subsubsection{\textbf{Decision-only}}
These attacks only require the final predicted labels. The \underline{q}uantum \underline{b}oundary \underline{a}ttack (QBA) adapts this by gradually shrinking a random perturbation \(\delta\) until model flips the label \cite{brendel2017decision,liao2021robust,west2023benchmarking}.
\begin{algorithm}[h]
\caption{QBA Pseudo Code.}
\KwIn{Initial state: $\rho$, function: $f$, step size: $\alpha_0$, decay factor: $\tau<1$, and total iterations: $K$.}
\KwOut{$\rho + \delta_K$.}

Initialize $\delta_0$ so that $f(\rho + \delta_0) \neq f(\rho)$\;

\For{$k = 0$ \KwTo $K$}{
    Sample random direction $u_k$ from the unit sphere\;
    $\delta' \leftarrow \delta_k + \alpha_k\cdot u_k$\;
    
    \If{$f(\rho + \delta') \neq f(\rho)$ \textbf{and} $\|\delta'\| < \|\delta_k\|$}{
        $\delta_{k+1} \leftarrow \delta'$\;
    }
    \Else{
        $\alpha_{k+1} \leftarrow \tau\cdot\alpha_k$ \quad (\text{with } $\tau<1$)\;
    }
}
\Return{$\rho + \delta_K$}\;
\label{algo:boundary-attack}
\end{algorithm}

This attack converges in \(O(\sqrt{d})\) queries, where \(d\) is the Hilbert‐space dimension. Enhancements that sample from gradient‐informed subspaces (via finite differences) can decrease convergence time \cite{brendel2017decision,liao2021robust}. Algorithm \ref{algo:boundary-attack} shows how boundary attack works.
\subsubsection{\textbf{Model Extraction and IP Theft}}
A QML model consists of a variety of intellectual properties (IPs) including the number of parameters and the layer depth. Illegal access to the components of the model such as architecture, training data, and encoding techniques is known as model extraction and IP theft \cite{10.1145/3696843.3696846,kundu2024evaluating}.

{Oracle-based attacks provide a more realistic threat model by assuming limited access to the target system through queries. These approaches are relatively less explored in the QAML literature compared to white-box attack settings. Existing studies often rely on simplified query mechanisms or surrogate models, and their evaluation is frequently restricted to simulation environments. This gap suggests that further investigation is needed to understand the effectiveness and limitations of query-based attacks in practical QML deployments.}
\subsection{\textbf{Circuit-level and Hardware-level Attacks}}
Attacks on VQCs take advantage of direct access to the circuit's parameters, either during training or at inference, to push the circuit into adversarial regimes \cite{9144562,10528776}. Studying these attacks is essential, since they strike at the core learning mechanism of QML, bypass data‐level protections, and show how circuit design, ansatz depth, and noise together influence the robustness \cite{Arias2023Survey,franco2024predominant}.
\subsubsection{\textbf{Fault Injection}}
A primary threat on this surface is fault injection, where an adversary maliciously alters the QML model's architecture/parameters before or during the deployment. The most insidious example of this is the \underline{Quantum Trojan}, a hidden backdoor secretly embedded into a QNN's structure. This trojan can remain dormant during normal operation, only to be activated by a specific trigger, giving the attacker control over the model's output with a very high success rate \cite{franco2024predominant,das2023trojannet,upadhyay2024stealthy}. Quantum poison (Q-Poison) \cite{ergu2025q} stealthily manipulates the parameters of quantum gates in hybrid quantum-classical networks (HQCNNs) before model's deployment to cause misclassification during the test phase, while maintaining the accuracy during training and validation.
\subsubsection{\textbf{Software Stack Exploitation}}
Another significant vector for circuit-level attacks is the exploitation of the quantum software stack. Adversaries can target vulnerabilities inherited from the classical languages that QML frameworks are built upon (e.g., Python in Qiskit) or find and exploit bugs within the quantum compilers and libraries themselves \cite{Arias2023Survey}.

\subsubsection{\textbf{Anti Noise‐Injection}}
Some defenses add random perturbations to parameters, \(\theta_i \to \theta_i + \delta_i\) with \(\delta_i\sim\mathcal{N}(0,\sigma^2)\), where $\mathcal{N}(0,\sigma^2)$ is a normal Gaussian distribution with mean zero and standard deviation $\sigma$. Attackers respond by optimizing the expected loss under this noise,
\begin{equation}
\min_{\delta}\; \mathbb{E}_{\delta_i\sim\mathcal{N}(0,\sigma^2)}\bigl[\ell(\boldsymbol{\theta} + \delta + \delta_i)\bigr]  
\end{equation}

This is usually approximated by sampling \(B\) noise realizations per update \cite{franco2024predominant}.  Randomized compiling, which injects random Pauli gates to scramble errors, has also been adapted to break up adversarial interference patterns.

In practice, attackers solve:
\begin{equation}
\max_{\|\delta\|\le\varepsilon}\;\frac{1}{B}\sum_{b=1}^B \ell\bigl(\boldsymbol{\theta} + \delta + \delta^{(b)}\bigr)    
\end{equation}
using batch gradient estimates, so their perturbation works across defense randomness.  With \(B\) noise samples per step, randomized defenses can partially recover attack success; exact recovery rates vary by circuit and noise model.
\subsubsection{\textbf{Hardware Specific}}
The broadest and most physically invasive attack surface is the quantum hardware itself, including the Quantum Processing Unit (QPU) and its extensive classical control infrastructure \cite{Arias2023Survey, franco2024predominant, Xu2023Classification}. These attacks go far beyond the digital realm, including novel quantum-specific fault models like the malicious manipulation of the analog microwave pulses used to control qubits \cite{Xu2023Classification}.

Practical examples of such threats are particularly relevant in shared quantum cloud computing environments. A malicious co-tenant could execute a carefully designed quantum program to intentionally induce hardware errors that affect a victim's circuit running on the same processor. These attacks could include generating \underline{cross-talk} between adjacent qubits on a superconducting chip or manipulating the \underline{ion shuttle operations} on a trapped-ion processor, leading to performance degradation or a denial-of-service \cite{franco2024predominant,saki2021shuttle,upadhyay2022robust,ghosh2025primersecurityquantumcomputing}. The successful implementation of QML models on real superconducting and trapped-ion processors underscores the real-world relevance of securing these physical systems against such threats \cite{Ren2022Experimental, Zhang2025Experimental,Jin2025Realizing}.

Defenses that operate on this surface include logic obfuscation to protect circuit IP \cite{das2023randomized,upadhyay2024obfuscating}, the insertion of trainable noise layers to enhance robustness \cite{huang2023enhancing}, and the application of randomized encoding circuits that mask gradient information from the attacker \cite{gong2024randomized}. These defenses highlight the criticality of the circuit's logical structure as a valuable and attackable asset.

{Circuit-level and hardware-level attacks extend the adversarial surface beyond input manipulation by targeting both the logical structure and the physical implementation of quantum systems. These attacks highlight important risks related to model integrity, execution reliability, and potential leakage of sensitive information. However, their feasibility often depends on the level of access to circuit design, compilation workflows, or hardware control mechanisms, which may vary across deployment settings. In addition, their impact is closely tied to system-specific factors such as circuit architecture, optimization procedures, and device characteristics, making it difficult to generalize findings across different platforms. As a result, while these attacks reveal critical vulnerabilities in the broader QML stack, their practical scope and consistency across real-world scenarios remain to be systematically established.}
\subsection{\textbf{Application‐Specific Adversarial Methods}}
Generic attacks target abstract QML models, but real applications use specialized encodings and cost functions that bring their own risks. Here, we review three key areas, quantum image recognition, quantum NLP with kernel SVMs, and variational algorithms for chemistry and finance, showing how each domain’s representations affect adversarial strength.
\subsubsection{\textbf{Quantum Image Recognition}}
Quantum convolutional neural networks (QCNNs) map an \(N\)-pixel image \(\mathbf{x}=(x_1,\dots,x_N)\) into a quantum state, either by amplitude encoding,
\begin{equation}
|\rho_{\mathrm{img}}\rangle
= \sum_{i=1}^N \sqrt{\frac{x_i}{\|\mathbf{x}\|_1}}\,|i\rangle,    
\end{equation}
or by angle encoding with \(R_Y(x_i)\) rotations on individual qubits.  Attackers adapt FGSM and PGD to perturb either amplitudes or angles:
\begin{align}
&\rho' = \mathrm{FGSM}_\varepsilon(\rho),\\
&\rho_{k+1} = \Pi_{\mathcal{B}}\bigl(\rho_k - \alpha\,\nabla_{\rho}\ell(\rho_k)\bigr),
\end{align}
and report a high misclassification rates on malware detection and handwritten‐digit tasks with 8-qubit simulators \cite{akter2023exploring}.  Since amplitude encoding spreads small pixel changes across all qubits, QCNNs are especially vulnerable to global tweaks, while angle encoding limits the effect locally.

Clipping pixel amplitudes to \([x_{\min},x_{\max}]\) can decrease attack success. Adding randomized phase shifts \(R_Z(\phi)\) after encoding can further reduce transferability.
\subsubsection{\textbf{Quantum NLP and Kernel SVMs}}
In Quantum NLP, text tokens \(t\) are turned into quantum feature maps \(\phi(t)\in\mathcal{H}\) (Hilbert space), via parameterized circuits.  A kernel
\begin{equation}
K\bigl(\phi(t_i),\phi(t_j)\bigr)
= \bigl|\langle\phi(t_i)|\phi(t_j)\rangle\bigr|^2 
\end{equation}
is plugged into a classical SVM. Adversaries apply small rotations in the feature space,
\begin{equation}
|\phi'(t)\rangle
= U(\delta)\,|\phi(t)\rangle,\quad \|\delta\|\le\varepsilon,    
\end{equation}
where $U$ is a unitary operation (matrix). This process shifts the kernel matrix and the SVM's decision boundary. Poisoning during the training session, which includes injecting crafted token sequences with incorrect labels, can significantly drop the accuracy of sentiment analysis tasks on standard benchmarks. This happens due to the kernel’s sensitivity to outliers \cite{montalbano2025quantum,franco2024predominant}.

Adding \(\lambda\|K-W\|_F^2\) to the SVM loss, where \(W\) is a smoothed kernel, can reduce the poisoning effect for \(\lambda=0.1\), with only a small increase in the training time.

{Table \ref{tab:attack-research-works} presents a comprehensive overview of adversarial attack studies on QML models. It shows that adversarial research in QML is still at an early and limited stage, with most studies focusing on evasion attacks under strong white-box assumptions. Moreover, these approaches are typically evaluated on small-scale simulated platforms using simple benchmark datasets such as MNIST.}
\begin{table*}[htbp]
    \centering
    \caption{Most significant adversarial attack studies on QML models - perturbation constraint (PC), Number of Qubits (NoQ), circuit depth (CD), Medical MNIST (MMNIST), Fashion MNIST (FMNIST), Kuzushiji MNIST (KMNIST), hardware and circuit level (HCL), and Hilbert-Schmidt (HS).}
    \resizebox{\linewidth}{!}{
    \begin{tabular}{l c c c c c c c c c}
        \hline
        Ref & Type & \multicolumn{3}{c}{Knowledge} &\makecell{PC}&NoQ&\makecell{CD}&Dataset(s)&Platform\\
         &&\rotatebox{270}{white}&\rotatebox{270}{black}&\rotatebox{270}{gray}&&&&&\\ 
         \hline
         \cite{lu2020quantum}&Evasion&\checkmark&\checkmark&\ding{55}&$\ell_p$&$8-10$&$\leq$50&MNIST, Ising&Yao.jl, Flux.jl, Zygote.jl\\
         \cite{akter2024quantum}&Evasion&\checkmark&\ding{55}&\ding{55}&$\ell_\infty$&2&N/A&ClaMP&Pennylane\\
         
         \cite{west2023benchmarking}&Evasion&\checkmark&\checkmark&\ding{55}&$\ell_\infty$&$10-12$&$200-1000$&MNIST, CIFAR-10, Celeb-A&Pennylane\\
         
         \cite{gong2022universal}&Evasion&\checkmark&\ding{55}&\checkmark&HS ($\epsilon$-bounded)&8&$5-10$&MNIST, Ising&Real Hardware\\
         
         \cite{el2025designing}&Evasion&\checkmark&\ding{55}&\ding{55}&$\ell_\infty$&4&N/A&MNIST, FMNIST&N/A\\
         
         \cite{qiu2023universal}&Evasion&\ding{55}&\checkmark&\ding{55}&$\ell_\infty$&12&20&MNIST, MMNIST&Yao.jl\\

         \cite{Li2025Computable}&Evasion&\checkmark&\ding{55}&\ding{55}&$\ell_2$&10&200&MNIST, FMNIST&Pennylane\\
         
         \cite{Wu2023Radio}&Evasion&\checkmark&\ding{55}&\ding{55}&$\ell_\infty$&8&30&RML, Synthetic&Pennylane\\

         \cite{zhao2025black}&Poisoning&\ding{55}&\checkmark&\ding{55}&$\ell_\infty$&10&20&MNIST&Pennylane\\
        
         \cite{huang2023backdoor}&Poisoning&\ding{55}&\checkmark&\ding{55}&$\ell_\infty$&$6-8$&$6-19$&MNIST&Pennylane\\
         
         \cite{kundu2025adversarial}&Poisoning&\ding{55}&\ding{55}&\checkmark&None&4, 8&2&MNIST, FMNIST, KMNIST, Letters&Pennylane\\
         
         \cite{chen2026superior}&Poisoning&\checkmark&\ding{55}&\ding{55}&None&10, 12&4, 6&MNIST, XXZ&TensorCircuit\\
         
         \cite{ghosh2024quantum}&Oracle&\ding{55}&\ding{55}&\checkmark&None&$1-8$&$1-3$&MNIST&Qiskit, Pennylane\\

         \cite{Wendlinger2024Comparative}&Oracle&\ding{55}&\ding{55}&\checkmark&$\ell_\infty$&8&32&Synthetic&Pennylane\\
         \cite{Ghosh2024Reverse}&Oracle&\ding{55}&\ding{55}&\checkmark&None&$1-8$&$1-3$&MNIST&Qiskit\\

         \cite{kundu2024evaluating}&Oracle&\ding{55}&\checkmark&\ding{55}&None&$2-8$&$2-16$&MNIST, FMNIST, KMNIST, Letters&Pennylane\\
         
         \cite{fu2024quantumleak}&Oracle&\ding{55}&\checkmark&\ding{55}&None&4&2&MNIST, FMNIST&Qiskit\\
         
         \cite{Wu2023Radio}&Oracle&\ding{55}&\checkmark&\ding{55}&$\ell_2$&8&30&RML, Synthetic&Pennylane\\
         \cite{upadhyay2025quantum}&Oracle&\ding{55}&\ding{55}&\checkmark&None&$3-14$&$\sim 67 - 95$&Synthetic&Qiskit\\
         
         \cite{kundu2025inverse}&Oracle&\ding{55}&\ding{55}&\checkmark&None&$4-12$&50&Synthetic&Qiskit\\
         
         \cite{choudhury2024crosstalk}&HCL&\ding{55}&\ding{55}&\checkmark&None&$7-16$&N/A&MQTBench&Real Hardware\\
         
         \cite{chu2023qtrojan}&HCL&\ding{55}&\ding{55}&\checkmark&None&4, 16&4&MNIST, sin functions&Qiskit\\
         
         \cite{ergu2025q}&HCL&\checkmark&\ding{55}&\ding{55}&$\ell_\infty$&N/A&N/A&Synthetic&Pennylane\\
         
         \cite{xu2025security}&HCL&\ding{55}&\ding{55}&\checkmark&None&2, 4&N/A&MNIST&Qiskit\\

         \cite{chu2023qdoor}&HCL&\checkmark&\ding{55}&\ding{55}&None&2, 8&N/A&MNIST, FMNIST, IRIS&Qiskit, BQSKit\\

         \cite{montalbano2025quantum}&App Specific&\checkmark&\ding{55}&\ding{55}&$\ell_2$&10&77&MMNIST&Qiskit\\

         \cite{akter2023exploring}&App Specific&\ding{55}&\checkmark&\ding{55}&None&N/A&2&ClaMP&Pennylane\\
         \hline
    \end{tabular}}
    \label{tab:attack-research-works}
\end{table*}
\section{\textbf{Defense Strategies in Quantum Settings}}
\label{sec:defenses-in-qml}
The inherent susceptibility of CML models to adversarial perturbations makes QML models similarly vulnerable to these attacks. As quantum computing capabilities advance and introduce new security risks, many architecture-agnostic classical defenses can be used and extended to the QML settings. Additionally, studying evolving threats, such as gradient-free attacks, has directed research toward quantum-specific protection mechanisms \cite{west2023towards,kejriwal2025advancing}.

\begin{figure}[htbp]
    \centering
    \includegraphics[width=0.8\linewidth]{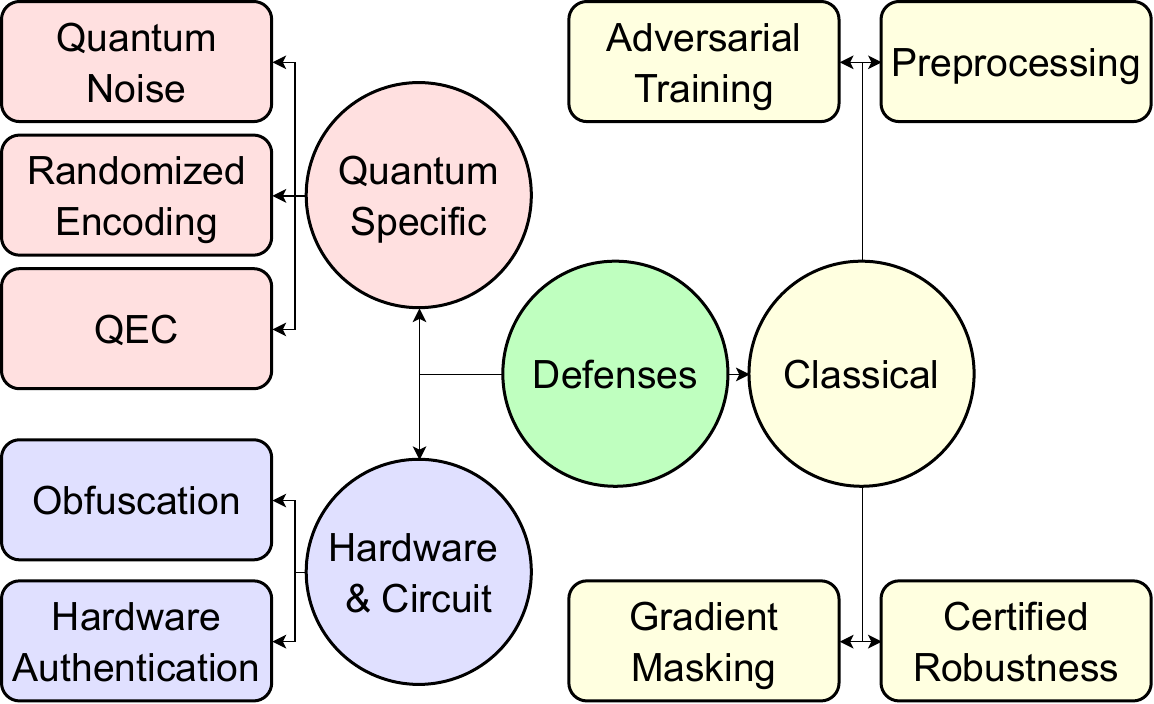}
    \caption{Major defense types in QAML - quantum error correction (QEC).}
    \label{fig:defense-taxonomy}
\end{figure}

The integration of quantum computing with ML introduces unique security concerns beyond traditional ML threats, including vulnerabilities in quantum classifiers with high-dimensional Hilbert space and attacks that exploit quantum data encodings or noise. Given the growing practical potential and complexity of QML, the necessity of novel strategies to maintain its security without sacrificing quantum advantage is reinforced \cite{franco2024predominant,yocam2024quantum}. Figure \ref{fig:defense-taxonomy} shows a taxonomy of defense strategies for QML models.

\subsection{\textbf{Classical Defense Adaptations}}
Like classical models, QML models remain susceptible to adversarial attacks due to inherited vulnerabilities. Researchers have shown that many classical defense techniques, due to their architecture-agnostic nature, such as adversarial training, gradient masking, preprocessing, and certified techniques, can be implemented in QML frameworks. This demonstrates the transferability between classical and quantum domains, potentially improving models' robustness against adversarial threats \cite{west2023towards,yocam2024quantum}.
\subsubsection{\textbf{Adversarial training}}
Adversarial training has emerged as a fundamental mechanism against attacks in both CML and QML models. In a quantum setting, this approach trains a QML model using both benign data and adversarially perturbed samples. These adversarial inputs contain subtle perturbation designed to cause misclassification, while remaining virtually imperceptible to human observers, which enhance the model's resilience against the adversaries \cite{franco2024predominant,west2023towards,lu2020quantum,akter2024quantum,marchiori2025attaq}. Early experiments highlight the susceptibility of QML to well-known classical attacks, such as FGSM, BIM, PGD, and Momentum Iterative Method (MIM) \cite{dong2018boosting}. Adversarial training can {improve robustness against certain classes of attacks, particularly those used during training} \cite{franco2024predominant}.

As discussed in \cite{lu2020quantum}, this process can be formulated as a robust min-max optimization task:
\begin{equation}
\min_{\theta} \frac{1}{N} \sum_{i=1}^N \max_{U_\delta \in \Delta} \mathcal{L}(h(U_\delta|\rho^{(i)}_\text{in},\theta),y^{(i)})
\label{eq:1}
\end{equation}
where $\lvert \rho_{\text{in}}^{(i)} \rangle$ is the $i$-th quantum input state, $y^{(i)}$ is its label, and $\delta$ is the adversarial perturbation. The formulation of equation \ref{eq:1}, which is a min-max optimization problem, is well studied in robust optimization, with established solution methods. As an efficient method, we break the problem into two parts. The inner maximization finds the worst-case adversarial perturbation, and the outer minimization updates the parameters of the model to reduce the loss on adversarial examples \cite{lu2020quantum}.
 
In \cite{Ren2022Experimental}, adversarial training was successfully demonstrated on a 10-qubit superconducting quantum processor. The authors trained variational quantum classifiers on both classical and quantum datasets, and showed that generating adversarial examples during training significantly improved robustness. The adversarially retrained classifier was able to correctly classify previously misclassified inputs, achieving near-perfect robust accuracy even in the presence of adversarial noise. 

Similarly, in \cite{montalbano2025quantum}, the authors explored quantum kernel methods and demonstrated that hybrid quantum kernel SVM classifiers are vulnerable to adversarial examples. They showed that augmenting training data with adversarial samples significantly boosted the robustness of these classifiers.
\subsubsection{\textbf{Gradient masking}}
Gradient masking is a classical adversarial defense method, in which the gradients of the model's loss function are intentionally obscured or rendered uninformative to hinder gradient-based adversarial attacks, particularly in white-box settings \cite{lu2020quantum,athalye2018obfuscated}. In \cite{li2024privacy}, the authors aim to prevent data leakage in a federated setting by masking the gradient information inside quantum phase and amplitude encodings, so that only aggregated values can be recovered rather than individual client's gradients.

The literature on classical adversarial
ML has shown that many gradient-masking defenses can be bypassed by adaptive attacks \cite{athalye2018obfuscated}. Quantum researchers remain cautious for the same reason, masked gradients might stop straightforward quantum-gradient-based attacks, but more clever strategies  could overcome this type of defense. Additionally, in black-box attacks, where the attacker has limited or no information about the model's internal structure, this method becomes less effective. While gradient masking techniques have been explored in quantum adversarial contexts, they are generally viewed as incomplete defenses that should be combined with other methods to be considered as a reliable defense mechanism \cite{gong2024randomized,lu2020quantum,athalye2018obfuscated}.
\subsubsection{\textbf{Preprocessing}}
Preprocessing the inputs to remove or limit adversarial perturbations is another classical idea that has been applied in QML settings. Input transformations such as, reducing image resolution, converting to grayscale, filtering noise, or data compression, can potentially shrink the space in which adversarial perturbations live. In the study by Ren et al., the authors converted all classical MRI images to 16×16 grayscale before encoding them into the quantum circuit \cite{Ren2022Experimental}. This drastic dimensionality reduction was primarily to meet the constraints of their quantum hardware (limited qubit count and circuit depth), but it faced a side-effect, where much of the high-frequency detail was removed. High-frequency image details are often where adversarial noise hides; therefore, such preprocessing could incidentally make it harder for an adversary to craft a subtle perturbation that survives the encoding. In essence, a coarse input encoding acts similarly to a denoising filter or compression defense, potentially reducing the “effective” adversarial search space \cite{zhang2024quantum}. 

Das et al. \cite{das2017keeping} found that retraining a model on JPEG-compressed CIFAR-10 images significantly boosted its accuracy on FGSM adversarial examples (from ~29\% to ~80\%). The compression’s removal of high-frequency noise (via the discrete cosine transform) likely accounts for this improvement \cite{yocam2024quantum}. By analogy, quantum classifiers might also benefit from classical preprocessing steps like compression or denoising before data is fed into the quantum circuit. As experiment has been performed in \cite{Ren2022Experimental}, which already supports this idea by using low-resolution inputs. 

It’s conceivable that explicit denoising algorithms (e.g. blurring, median filters) or dimensionality reduction (e.g. PCA) could be applied on classical data prior to quantum encoding to mitigate adversarial perturbations, though the research on this area is still scarce. Another related approach is to remove suspicious samples from the training data before training the model. Quantum detection (Q-Detection) \cite{he2025q} identifies poisoned samples by solving a bilevel meta-optimization problem and removes them before training.
\subsubsection{\textbf{Certified Robustness}}
These defenses simulate the worst case scenario of the attacks, and aim to robustify the network using different verification techniques such Lipschitz continuity and differential privacy \cite{franco2024predominant,du2021quantum,watkins2023quantum}. These verification techniques are discussed in more detail in Section \ref{sec:theoretical-perspectives}.

{Defense strategies such as adversarial training, gradient masking, and preprocessing techniques extend established approaches from classical machine learning into the quantum domain. These methods aim to improve robustness by exposing models to adversarial variations or by transforming inputs prior to processing. However, their effectiveness can be sensitive to the choice of attack models used during training and may not generalize well to unseen or structurally different adversarial strategies. In addition, integrating these approaches into quantum pipelines introduces additional complexity due to encoding dependencies and circuit constraints. As a result, while these defenses offer a natural starting point, their robustness under diverse and adaptive attack scenarios remains uncertain.}

Overall, classical defense mechanisms remain a valuable toolkit in the quantum setting. Their effectiveness, however, must be rigorously validated under physically realizable quantum noise and hardware limitations.

\subsection{\textbf{Quantum-specific Defense Mechanisms}}
The eventual failure of the classical defense strategies against more sophisticated attacks has shifted the research to investigate defenses depending on quantum phenomena. Quantum-specific defense mechanisms use uniquely quantum properties to provide new avenues for securing QML models. Early studies established that quantum classifiers can be as vulnerable as classical ones. For instance, Liu and Wittek \cite{liu2020vulnerability} showed that the perturbation needed to fool a quantum model decreases as the Hilbert space dimension grows. This reveals a fundamental robustness versus complexity trade-off, with high-dimensional quantum models requiring additional resources (e.g. qubits or error mitigation) to preserve security. Such findings motivate defense strategies tailored to the quantum domain. These techniques include quantum noise, randomized encoding, and quantum error correction \cite{west2023towards}.
\subsubsection{\textbf{Quantum noise}}
Quantum noise has been viewed as an obstacle in quantum computation, particularly in near-term NISQ devices. Recent research explores its potential as a defensive asset. Inspired by classical computing applications where noise aids in privacy, signal enhancement, and resolution improvement, quantum noise is being studied for its security benefits in QML \cite{du2021quantum}. 

In particular, depolarization noise in quantum classifiers has been shown to improve robustness against adversarial perturbations. This resilience increases with noise intensity and aligns with the principles of \underline{q}uantum \underline{d}ifferential \underline{p}rivacy (QDP), offering model-agnostic protection independent of internal classifier structure. Unlike classical defenses that rely on architectural details, {quantum noise can contribute to robustness under certain conditions, although excessive noise may degrade overall performance} \cite{du2021quantum}.

Increasing quantum rotation noise during training can directly increase a classifier's robustness. In \cite{franco2024predominant}, the theoretical analysis of quantum rotation noise in adversarial defense demonstrates its role in enhancing classifiers' resilience similar to classical randomized smoothing. It is provable that increasing rotation noise during training directly improves robustness, with privacy guarantees inversely scaling with noise intensity. The previous study expanded by connecting \underline{q}uantum \underline{h}ypothesis \underline{t}esting (QHT) to gain robustness against unanticipated noise. Their work introduced certified robustness protocols, classification validation methods in a noisy environment, and conditions for robustness under different types of noise, all formalized via QHT.

Quantum noise injection for adversarial defense (QNAD) \cite{kundu2024qnad}, uses the inherent noise and crosstalk in quantum hardware as a defensive aid. By introducing a controlled amount of noise (analogous to a noisy input transformation), the network’s accuracy under adversarial environment improved significantly. This counterintuitive strategy uses quantum noise to essentially “denoise” the adversarial effect, the added randomness makes the model’s inference less sensitive to the precise perturbations crafted by the attacker, much like how adding random jitter or performing random input rotations can thwart classical adversarial attacks.

In summary, a modest level of quantum noise can act as a regularizer against adversarial perturbations, a concept distinct from but analogous to classical noise-injection defenses. While excessive noise may degrade accuracy, careful noise levels can significantly blunt an attack’s effectiveness. This noise-based defense has been validated in theory with bounds and simulations, hinting that inherent decoherence in NISQ devices might naturally provide some robustness \cite{du2021quantum}.
\subsubsection{\textbf{Randomized Encoding}}

Randomizing the data encoding process provides a powerful defense against adversarial threats in quantum learning, including gradient-based attacks and worst-case experimental noise. By implementing randomized unitary encoding, we can effectively mask critical gradient information from potential adversaries \cite{gong2024randomized,upadhyay2025quantum}.

By using encoders that satisfy the unitary 2-design property, the expected gradient values for adversarial variational circuits can be exponentially suppressed, leading to barren plateaus, where gradients vanish exponentially with system size. This phenomenon significantly impedes white-box adversaries relying on gradient calculations as shown in (\ref{eq:2}), to receive exponentially diminished values under random unitary encoding, effectively masking critical information \cite{gong2024randomized}:

\begin{equation}
    \max_{\delta \in \Delta} \mathcal{L}\left(h(U_\delta |\rho^{(i)}\rangle; \boldsymbol{\theta}), s^{(i)}\right)
    \label{eq:2}
\end{equation}
where $\delta$ is the adversarial perturbation and $U_\delta$ is a parametrized unitary perturbation. Theoretical guarantees show that unitary 2-designs reduce first and second-moment gradient expectations near zero, forcing attackers to require exponentially high precision and iterations. 

As demonstrated in \cite{gong2024randomized}, the probability of observing large gradients is bounded by $\mathcal{O}(1 / \delta d^2)$, where $d$ is dimension of the Hilbert space, via Chebyshev’s inequality, making successful attacks increasingly inefficient. Importantly, the encoding scheme requires only $\mathcal{O}(n^2)$ quantum gates, aligning with the scaling of practical quantum classifiers. Unlike prior methods that embedded randomness directly into the model, the barren plateaus in this framework arise from the encoder’s unitary 2-design property. This distinction provides strong protection without constraining the model architecture, offering a scalable and architecture-agnostic defense mechanism. 

Additionally, since the randomization does not compromise the legitimate classification (the model applies the inverse transformation before inference), the accuracy remains intact while security is enhanced. Such encoding-based defenses illustrate how secret unitaries or basis rotations can safeguard quantum models by injecting uncertainty and complexity that adversaries cannot account for \cite{gong2024randomized}.
\subsubsection{\textbf{Quantum Error Correction}}
Quantum error correction (QEC) plays an important role in protecting quantum information from errors, whether caused by noise or adversarial interference. By encoding a single logical qubit across multiple physical qubits, QEC allows for the detection and correction of localized errors without compromising the integrity of the logical qubit.  This redundancy strengthens the quantum system's resistance to both stochastic and malicious attacks \cite{yocam2024quantum,lenssen2025fooling,chatterjee2025q}. Gong et al. \cite{gong2024randomized} investigated error-correcting encoders as a defense, where input states are encoded into a larger Hilbert space using a QEC code before classification, so that any minor adversarial disturbance is actively corrected during decoding. Their analysis showed that random black-box QEC encoders can protect quantum classifiers against localized adversarial noise, where robustness improves as more layers of error-correcting codes are added.

QEC has shown several advantages, including enhanced protection against noise-induced attack vectors, improved stability and reliability of quantum computations, and ongoing theoretical advancements of error-correcting defense mechanisms \cite{yocam2024quantum}. QEC is foundational for ensuring the security and correctness of protocols against both noise and quantum adversaries. Fault-tolerant quantum cryptographic protocols use QEC to enhance resistance against quantum adversaries \cite{saiyed2025quantum}. Implementing QEC, however, introduces resource overhead due to the increased qubit counts and circuit complexity for encoding and decoding operations, as well as trade-offs in code efficiency and fault-tolerance thresholds \cite{yocam2024quantum}.

Moreover, black-box QEC encoders can effectively amplify the QDP property, significantly enhancing the robustness of quantum classifiers against local adversarial attacks. QEC demonstrates strong resilience against adversarial disturbances, even in worst-case noise scenarios. Some analysis reveals that the number of QEC levels needed scales double logarithmically with the number of qubits $n$, that is, $\log (\log (n))$, suggesting that fault-tolerant quantum computers using black-box QEC codes can efficiently defend against local adversarial attacks with high probability \cite{gong2024randomized}.

While implementing full QEC for machine learning is resource-intensive, these methods underscore a principle: robustness through redundancy, where the quantum model can tolerate or auto-correct certain perturbations, thereby nullifying adversarial attempts that fall below the code’s error threshold.

In summary, QAML defense research has expanded beyond porting classical techniques to explore quantum-native strategies. Noise-assisted training, randomized data encodings, and QEC-based protections each exploit different aspects of quantum physics to thwart adversaries. These mechanisms tend to smooth out or eliminate the sharp gradients that attackers rely on, whether by randomizing the model’s response or by correcting the perturbations outright. Although largely demonstrated in theory and small-scale experiments, such quantum-specific defenses illustrate a promising advantage: {quantum models possess unique structural and physical properties that may offer inherent robustness advantages in certain settings, although the extent and consistency of these benefits remain to be systematically validated.} As quantum hardware improves, these defense mechanisms are expected to play a crucial role in securing quantum machine learning systems against adversarial manipulation.

{Defenses based on noise injection, randomized encoding, or stochastic transformations introduce variability within the system to reduce the impact of adversarial perturbations and information leakage. These approaches can introduce uncertainty that disrupts structured attacks, but their effectiveness depends on how this variability interacts with the model. In some cases, increased randomness can also degrade predictive performance or stability, leading to trade-offs between robustness and accuracy. Additionally, such defenses may introduce computational and resource overhead, potentially increasing training and inference costs.}
\subsection{\textbf{Hardware-level and Circuit-level Defenses}}
Quantum adversaries may target the physical implementation of QML models, for instance by manipulating quantum gates, introducing tailored noise, or exploiting hardware backdoors. The protection at the hardware and circuit levels is important to counter threats originating from the infrastructure and operation of quantum systems.
\subsubsection{\textbf{Obfuscation}}
In near-term noisy quantum computers, shorter and simpler circuits are more reliable. While quantum compilers optimize circuits by translating high-level gates into hardware-compatible ones, using untrusted third-party compilers introduces security risks, including IP theft. A novel approach is randomized reversible gate-based obfuscation, which protects circuits during compilation. This method inserts a small random circuit and its inverse to disrupt the circuit’s functionality before compilation. Afterward, the inverse is appended to restore the original behavior, providing security with minimal fidelity loss ($1-3\%$) \cite{das2023randomized,upadhyay2024obfuscating}.

Obfuscation in quantum circuits is challenging due to the probabilistic nature of quantum outputs and narrow fidelity margins. For example, a circuit with a 95\% correct output must be obfuscated beyond that threshold to be effective; the obfuscation technique must overcome this 90\% reliability gap to effectively corrupt the output. To improve obfuscation, the randomly inserted circuit can be refined to maximize functionality disruption. Additionally, where the random circuit is placed, plays a critical role, as it affects both how well the output is corrupted and how easily the user can reverse the obfuscation later. Strategic placement at the front, middle, or back helps to maximize output corruption, leaving fewer clues for attackers and minimizes detection by untrusted compilers \cite{das2023randomized,ghosh2025primersecurityquantumcomputing,upadhyay2024trustworthy}. Such randomness-based defenses effectively decorrelate the circuit execution from any specific fixed attack pattern, forcing potential adversaries into a worst-case (random-guessing) scenario.
\begin{table*}[htbp]
    \centering
    \caption{Most significant adversarial defenses for QML models - number of qubits (NoQ), circuit depth (CD), Medical MNIST (MMNIST), Fashion MNIST, (FMNIST), breast cancer (BC), Gaussian Mixture (GM), hardware and circuit level (HCL), and interleaved block-encoding (IBE).}
    \resizebox{\linewidth}{!}{
    \begin{tabular}{l c c c c c c c}
        \hline
        Ref & \makecell{Defense\\Type} & \makecell{Attack\\Type} & NoQ & CD & \makecell{Encoding\\Strategy} & Dataset(s) & Platform\\
         \hline
         \cite{lu2020quantum}&Classical&Evasion&$8-10$&$\leq 50$&Amplitude&MNIST, Ising&Yao.jl, Flux.jl, Zygote.jl\\
         \cite{Ren2022Experimental}&Classical&Evasion&10&$\leq 60$&IBE&MNIST, MRI, Synthetic&Real Hardware\\
         
         \cite{montalbano2025quantum}&Classical&Evasion&10&77&Angle&MMNIST&Qiskit\\
         \cite{maouaki2025qfalquantumfederatedadversarial}&Classical&Evasion&6&N/A&Amplitude&MNIST&N/A\\
         
         \cite{west2023benchmarking}&Classical&Evasion&$10-12$&$200-1000$&Amplitude&MNIST, CIFAR-10, Celeb-A&Pennylane\\
         
         \cite{marchiori2025attaq}&Classical&Evasion&N/A&N/A&Angle&MNIST, FMNIST, CIFAR-10&TorchQuantum\\

         \cite{wang2025qsentry}&Classical&Poisoning&8&8&Amplitude&MNIST&Pennylane\\

         \cite{li2024privacy}&Classical&Oracle&N/A&N/A&Amplitude&N/A&Qiskit\\
         
         \cite{liao2021robust}&Quantum-specific&Evasion&N/A&N/A&Angle&N/A&N/A\\
         
         \cite{berberich2024training}&Quantum-specific&Evasion&3&3&Angle&Circle&Pennylane\\
         
         \cite{gong2024randomized}&Quantum-specific&Evasion&$4-14$&4&Randomized&Ising&N/A\\
         
         \cite{li2026dual}&Quantum-specific&Evasion&N/A&N/A&Angle&Circle, GM, MNIST, BC&Pennylane\\
         
         \cite{wollschlager2024discrete}&Quantum-specific&Evasion&N/A&N/A&Amplitude&MNIST, IMDB, Synthetic&Qiskit\\
         
         \cite{franco2024quadratic}&Quantum-specific&Evasion&20&6&Basis&IRIS, GunPoint&Pennylane\\
         
         \cite{upadhyay2025quantum}&HCL&Oracle&$3-14$&$\sim 67-95$&Amplitude, Angle&Synthetic&Qiskit\\
         
         \cite{huang2023enhancing}&HCL&Evasion&4, 8&10&Amplitude, Angle&MNIST, Ising, Synthetic&Pennylane\\
         
         \cite{liu2025loq}&HCL&Oracle&$4-12$&$5-16$&None&RevLib benchmark circuits&Qiskit\\
         
         \cite{wang2023qumos}&HCL&Oracle&4&N/A&Angle&MNIST&Qiskit, TorchQuantum\\
         
         \cite{xu2025security}&HCL&HCL&2, 4&N/A&None&MNIST&Qiskit\\
         
         \cite{zhou2025watermarking}&HCL&Oracle&16&$\sim 5$&Amplitude, Angle&MNIST&Qiskit\\
         
         \cite{roy2025watermarking}&HCL&Oracle&$\sim 3-6$&$25-77$&None&RevLib benchmark circuits&Qiskit\\
         \hline
    \end{tabular}}
    \label{tab:defense-research-works}
\end{table*}
\subsubsection{\textbf{Hardware Authentication Techniques}}
In parallel, researchers are exploring hardware-level authentication and integrity measures to ensure that the quantum device and its delivered computations have not been tampered with.

\textbf{Quantum Physical Unclonable Functions (QPUFs)} represent a major advancement in quantum security by generating tamper-proof, device-specific cryptographic fingerprints based on the inherent randomness of quantum mechanics. Successful implementations have been demonstrated on IBM's quantum hardware and Google's Cirq simulator.
This technology on IBM quantum hardware achieved approximately $50\%$ intra-hamming distance, showing an exceptional reliability of $100\%$ on Google's simulator and $95\%$ on IBM's quantum simulator, and maintains consistent performance despite real-world quantum challenges like error rates and decoherence \cite{phalak2021quantum,bathalapalli2025qpuf}.

The QPUF circuit utilizes a carefully selected set of fundamental quantum gates, such as Hadamard, CNOT, Pauli-X, and Ry, to exploit quantum principles like superposition and entanglement, enabling the creation of device-specific identity keys. QPUF circuit innovation is designed to run on IBM's quantum computers, provides enhanced hardware-level security for smart devices, and creates device-specific identity keys using quantum effects to be used as unique device fingerprints \cite{phalak2021quantum,bathalapalli2025qpuf}.

\textbf{Watermarking} is especially effective against IP theft and model extraction attacks, where the ownership of the target model could easily be identified using watermarks. This approach embeds identifiable information into the model to facilitate ownership, authenticity, and unauthorized copies identification \cite{zhou2025watermarking,roy2025watermarking}.

\textbf{Quantum hardware logic locking and circuit authentication} is another line of defense. Inspired by classical logic locking, schemes like enhanced locking for quantum IP (E-LoQ) embed secret keys into quantum circuits such that the correct operation of certain gates is conditional on an unseen key qubit’s state \cite{liu2025loq}. 

Without the correct key, which is only known to the defender, the quantum circuit yields scrambled or incorrect results, thereby preventing an attacker with full access to the hardware or the compiled circuit from gleaning the true functionality or repurposing the circuit. Such locked circuits and obfuscation techniques inject high entropy into the circuit’s behavior, making it infeasible for an adversary to identify or exploit critical quantum operations without the secret key \cite{liu2025loq}. 

Likewise, trap qubits have been proposed as a form of circuit-level tamper-evidence: the defender inserts ancillary qubits prepared in known states at strategic locations in the circuit, which do not affect the computation when untouched, but will expose malicious interference if measured or altered unexpectedly. This concept, originally used in verified blind quantum computing protocols, allows a client to detect deviations by checking the outcomes of these trap qubits \cite{barz2012demonstration}. For instance, Barz et al. \cite{barz2012demonstration} showed that by interspersing trap qubits and pre-calibrated measurements into a delegated quantum computation, any attempt by a malicious quantum server to deviate or learn the client’s secret computation would be revealed with high probability. In the context of QAML, trap qubits or analogous test sequences could be embedded to ensure a quantum inference or training routine has not been perturbed by an adversary at the hardware level.

Hardware-level and circuit-level defenses collectively aim to improve the robustness of QML models against low-level adversarial manipulation. They assume threat models ranging from untrusted compilation tools and cloud providers to malicious co-tenants or altered devices. By introducing randomness and cryptographic control into the quantum circuit execution, these methods make it extremely challenging for an adversary to predict, influence, or replicate the model’s behavior without detection. However, design challenges remain. Randomization and obfuscation inevitably incur resource overhead (additional gates, deeper circuits, or extra qubits), which is problematic given the noise and decoherence limits of NISQ-era devices. For example, inserting protective gates or trap qubits can increase circuit depth and error rates, potentially affecting the QML model’s accuracy if not carefully managed. Hardware PUF responses must be repeatable despite quantum noise, requiring error mitigation or calibration to avoid false negatives in authentication \cite{das2023randomized,phalak2021quantum,gong2024randomized}. 

{Table \ref{tab:defense-research-works} presents a comprehensive overview of the most prominent adversarial defense strategies proposed for QML models. Notably, the majority of existing defenses are designed to mitigate evasion attacks, with comparatively limited attention given to other threat models such as poisoning or oracle-based attacks. This suggests that current research in QML security is largely focused on inference-time robustness, while training-time and more sophisticated adversarial scenarios remain relatively underexplored.}

Another challenge is ensuring that these defenses themselves are not bypassed or exploited, for instance, an adversary might attempt to learn the randomization seed or key, or target the defense components \cite{das2023randomized,phalak2021quantum,gong2024randomized}. By embracing strategies from randomization to device authentication, future quantum learning systems can be designed with built-in resilience that hardens them against attacks at the very foundations of quantum computation. Table \ref{tab:adefense_comparison} illustrates a comparative analysis of the effectiveness and limitations of different {defense} types.
\begin{table*}[htbp] 
    \caption{Comparative table of defense methods: strengths and limitations.}
    \begin{tabularx}{\linewidth}{@{} p{0.1\columnwidth} >{\RaggedRight}p{0.2\columnwidth} >{\RaggedRight}X 
    >{\RaggedRight}X @{}} 
        \toprule
        \textbf{Category} & \textbf{Approach} & \textbf{ Effectiveness} &
        \textbf{Main Limitation(s)/Trade-off(s)}\\
        \midrule
        Classical & Adversarial Training \cite{west2023towards,lu2020quantum}& Known attacks& Unknown attacks, Standard accuracy drop\\
        \cmidrule{2-4}

        & Preprocessing \cite{yocam2024quantum} & Shrinking adversary's space, Data poisoning & Insufficient experimental evidence\\
        \cmidrule{2-4}
        & Gradient Masking \cite{lu2020quantum} & Basic attacks & Adaptive attacks, Black-box attacks, Incompleteness\\
        \midrule
        Quantum & Quantum Noise \cite{du2021quantum, franco2024predominant} &  Gradient-based attacks, Decrease information leakage & Hardware dependency, Noise-induced errors\\
        \cmidrule{2-4}

        & Randomized Encoding \cite{gong2024randomized} & Gradient-based attacks, Decrease information leakage  & Resource overhead\\
        \cmidrule{2-4}
        & QEC \cite{gong2024randomized, yocam2024quantum} & Stabilizing system, Noise protection & Resource overhead\\
        \midrule
        Hardware & Logic Locking \cite{liu2025loq} & IP Protection & Fidelity loss, Computation overhead\\
        \cmidrule{2-4}
        & QPUF \cite{phalak2021quantum,bathalapalli2025qpuf} & Authentication & Hardware limitation, Temporal variation\\
        \cmidrule{2-4}

        & Watermarking \cite{zhou2025watermarking} & IP protection & Standard accuracy drop\\\cmidrule{2-4}
        & Obfuscation \cite{das2023randomized,gong2024randomized} & IP protection & Resource overhead\\
        \bottomrule
    \end{tabularx}
\label{tab:adefense_comparison}
\end{table*}

\section{\textbf{Theoretical Perspectives}}
\label{sec:theoretical-perspectives}
To design robust systems and accurately predict the behavior of quantum models in adversarial scenarios, it is essential to understand the theoretical underpinnings of QAML. These adversarial scenarios require a comprehensive analysis of the fundamental limits of robustness, the computational complexity of potential attacks, and the development of formal security guarantees.
\subsection{\textbf{Complexity-theoretic Viewpoints}}
The complexity of training and securing QML models presents unique challenges, often linked to the fundamental properties of quantum systems.

\subsubsection{\textbf{Complexity-enhanced hardness assumptions}}
Training VQAs and QML models is challenging due to their highly complex, non-convex loss landscapes. The primary challenges are barren plateaus (BPs), where gradients vanish exponentially with system size, and the growth of poor local minima, which are associated with shallow circuits. This challenge is prominent in deep quantum circuits and severely hinders gradient-based optimization due to the quantum curse of dimensionality \cite{bagaev2025regularizing, dowling2024adversarial, west2023towards}.
Barren plateaus can be used defensively in QML to protect against adversarial attacks. By employing random unitary encoders from a unitary 2-design, adversarial variational circuits can be forced into barren plateaus, making gradient extraction difficult. Theoretical analysis confirms that the gradient variance for any adversarial circuit parameter is bounded by an exponentially small value \cite{bagaev2025regularizing, west2023towards}. 

Chebyshev’s inequality further guarantees that the probability of finding a meaningful gradient is exponentially low, as gradients are overwhelmingly likely to be negligible. This makes it practically infeasible for attackers to extract meaningful gradient signals. Unlike prior findings showing vulnerability to perturbations scaling as $\mathcal{O}(1/d)$, with $d$ as the dimension of the Hilbert space, this method induces even faster gradient decay. 
The random encoder with $n$ qubits can be efficiently implemented with only $\mathcal{O}(n^2)$ gates. Importantly, this defense differs from traditional barren plateau studies. Here, the adversarial circuit remains near the identity operator and does not form a unitary 2-design; instead, the barren plateau arises solely from the random encoder. This approach effectively protects quantum learning systems by obscuring gradient information, forcing attackers to require exponential precision and iterations to construct adversarial examples \cite{gong2024randomized}.

\subsubsection{\textbf{Quantum-enhanced hardness assumptions}}

The rapid advancements of quantum computers spark research into their advantages for QML. Compared to CML, the advantage of QML may not necessarily lie in speed or accuracy, but in resiliency against adversarial attacks. As a result of differences in architecture, on standard image datasets,  quantum models can use features that are classically intractable to compute or exploit, which can provide an additional layer of robustness \cite{west2023benchmarking,dowling2024adversarial}. These findings highlight a natural expansion of classical notions of “robust” and “non-robust” features to include a third category, \reflectbox{"}classically intractable features". While these features are not guaranteed to be inherently robust, they are often resistant to classical adversarial strategies in practice \cite{west2023benchmarking}. The complexity lies in how quantum models perceive data differently. Classical adversarial examples, which are crafted to exploit features that classical networks rely on, tend to be less effective at fooling quantum models that do not depend on those same features. This means adversarial examples generated for classical models often fail to transfer effectively to quantum models, indicating a distinct but context-dependent quantum advantage in robustness \cite{dowling2024adversarial,bagaev2025regularizing}.

\subsection{\textbf{Robustness Bounds}}
Robustness in QML can be formally characterized through perturbation bounds and connections to frameworks. Understanding the limits and guarantees of robustness requires analyzing the sensitivity of quantum models to input perturbations and quantifying this behavior through rigorous mathematical bounds.
\subsubsection{\textbf{Lipschitz continuity and perturbation bounds in QML}}
Lipschitz continuity is a central theoretical tool for quantifying adversarial robustness and generalization in QML models. Perturbation bounds can be derived from Lipschitz, showing the sensitivity of a quantum model to input variations.
The Lipschitz continuity for QML models with trainable data encodings becomes parameter-dependent, indicating that the encoding norm affects robustness against data perturbations. By regularizing the Lipschitz bound during training, models can be made more robust and generalizable. This suggests that models with trainable encodings are crucial for systematically adapting robustness during training, as opposed to fixed encodings, where the Lipschitz bound cannot be influenced. The findings demonstrate that trainable encodings are essential for optimizing robustness and generalization in QML, offering both theoretical insights and practical training methods. Numerical experiments confirm the effectiveness of the proposed regularization strategy \cite{berberich2024training}.

On the other hand, in \cite{lu2020quantum}, the authors suggest that even small perturbations (scaling inversely with Hilbert space dimension) can cause quantum classifiers to misclassify inputs. This shows that robustness could diminish as quantum advantages grow in high-dimensional problems.
In \cite{west2023towards}, the authors highlight the concentration of measures in high-dimensional Hilbert spaces, for a system of $n$-qubits, where random inputs are extremely close to adversarial examples, with vulnerability scaling as $\mathcal{O}(2^{-n})$.
Universal adversarial examples can fool multiple classifiers in a $k$-class setting using perturbations as small as $\mathcal{O}(\log(k) \cdot 2^{-n})$.
For common encodings, the required perturbation strength may scale as $\mathcal{O}(1/\sqrt{n})$.

These results show the dual nature of QML robustness. While offering novel defenses, their high-dimensional structure introduces unique vulnerabilities. The incorporation of Lipschitz-based regulation balances these trade-offs.   

\subsubsection{\textbf{Quantum adversarial bounds and robustness analysis}}
QHT and QDP have been proposed to provide strong robustness bounds in different adversarial scenarios.

\textbf{Quantum Hypothesis Testing}: In quantum machine learning, ensuring the model's robustness against adversarial attacks and noise is a significant challenge. In \cite{weber2021optimal}, the authors established a fundamental and formal link between QHT and provably robust quantum classification. This connection provides a tight robustness condition, which is a precise measure of the amount of noise that a classifier can tolerate. This condition is crucial because it enables an accurate comparison of different quantum classifiers in terms of resilience, which was previously a major open problem. 

The QHT condition can be formulated as a semi-definite program (SDP), but the research also provides simpler, closed-form solutions based on key quantum metrics like Uhlmann fidelity, Bures metric, and trace distance. These solutions are proven to be sufficient and necessary for guaranteeing robustness in certain regimes, meaning they are both precise and reliable \cite{weber2021optimal}. By applying this framework, researchers can develop practical protocols to certify the model robustness against input perturbations and verify if the prediction of a noisy input is consistent with its unperturbed version.    

\textbf{Quantum Differential Privacy}: Adding depolarization noise to quantum circuits can be a strategic approach to enhance the robustness of QML models. While noise is typically seen as a flaw in quantum computing, this research shows it can be deliberately used to protect a quantum classifier against adversarial attacks. A key finding is that this noise-induced robustness provides a theoretical security bound that is independent of the model's specific classification details, which only depends on the number of output classes. This is a significant quantum advantage over classical protocols, where robustness bounds often depend on model specifics, such as the degree of nonlinearity \cite{du2021quantum}.

This method provides a strong, provable defense against even worst-case attacks and offers a unique way to use quantum noise beneficially. It also suggests that QML could provide a distinct type of advantage beyond just speed, offering a new avenue for securing both quantum and classical data. The research also opens up opportunities to explore the trade-offs between a model's robustness and its accuracy \cite{du2021quantum}.
\subsection{\textbf{Quantum PAC Learning and Security Definitions}}
Formal models and definitions provide a rigorous framework for understanding security and learning capabilities in the quantum domain.
\subsubsection{\textbf{Formal models for robust and secure QML}}
Quantum probably approximately correct (PAC) learning is a formal framework that measures the complexity of learning in quantum systems by counting the number of quantum state copies used. This model helps us understand the limits of quantum learning and its implications for cryptography and computation \cite{salmon2024provable}. 

While early research suggested only constant-factor advantages for quantum PAC learners over classical ones in general cases, recent work demonstrates a square root advantage (up to polylogarithmic factors) for quantum PAC learning in the full, general model. This highlights a generic quantum advantage in learning complexity. The Quantum PAC (QPAC) model's influence extends beyond core learnability theory, providing a foundational tool for analyzing the theoretical limits of other tasks in quantum computation and cryptography. For instance, the framework is instrumental in explaining the formal relationship between QDP and state learnability, as well as in assessing the fundamental complexity of training parameterized quantum circuits \cite{salmon2024provable}.

\subsubsection{\textbf{Security guarantees}}
Quantum classifiers possess inherent theoretical protections against classical adversaries. Specifically, they are proven to be: (i) robust to weak perturbations of data from the trained distribution, (ii) resistant to local attacks if they are insufficiently scrambling, and (iii) protected from universal adversarial attacks if they are sufficiently quantum chaotic. These analytic results, which are supported by numerical evidence, demonstrate the practical robustness of quantum classifiers. 
These guarantees rely on genuinely quantum properties of the QML architecture, such as the unitarity of quantum circuits, the dynamic complexity of the trained circuit, and the data encoding strategy. Chaotic unitaries effectively scramble information, making it difficult for an adversary to precisely manipulate the system to induce misclassification, thereby conferring robustness. Certain QML architectures (e.g., quanvolutional networks and quantum-enhanced RBMs) demonstrate inherent resilience towards adversarial attacks \cite{franco2024predominant,dowling2024adversarial}. 

Overparametrized QML models with highly expressive encodings offer inherent protection against gradient inversion attacks. In a federated learning context, theoretical and empirical evidence shows a perpetrator faces considerable challenges in crafting attacks on these models due to their complexity. The overparameterization of the trainable PQCs also ensures a training process free from spurious local minima \cite{franco2024predominant}.

{While these theoretical frameworks provide important insights about the behavior of QML models under adversarial environments, their practical specifications are also important. For instance, Lipschitz-based bounds can guide the design of quantum circuits with controlled sensitivity to input perturbations. QDP offers an approach to introduce noise that balances robustness and accuracy. Moreover, theoretical analyses based on QHT can facilitate the development of detection mechanisms and robustness evaluation protocols. Bridging these theoretical insights with practical model design and experimental settings is necessary to develop reliable and trustworthy QML systems.} Table \ref{tab:theoretical_comparison} reports a summary on each theoretical aspect and their guarantees.
\begin{table*}[htbp]
    \caption{Comparative table of theoretical results and guarantees.}
    \label{tab:theoretical_comparison}
    
    \begin{tabularx}{\linewidth}{@{} p{0.25\columnwidth} >{\RaggedRight}p{0.2\columnwidth} >{\RaggedRight}X @{}}
        \toprule
        \textbf{Theoretical Aspects} & \textbf{Key Concepts} & \textbf{Notes} \\
        \midrule
        \multirow{3}{*}{\makecell[l]{Complexity\\ \cite{west2023benchmarking,bagaev2025regularizing, dowling2024adversarial, west2023towards,gong2024randomized}}}& NP-hardness& Caused by non-convex landscapes and local minima.\\
        \cmidrule{2-3}
        & Barren Plateaus & Hindering trainability by causing vanishing gradients.\\
        \cmidrule{2-3}
        & Quantum-enhanced Hardness & More  resistant to classical attacks by using \reflectbox{"}classically intractable features".\\
        \midrule
        \multirow{3}{*}{\makecell[l]{Robustness Bounds\\ \cite{west2023towards,lu2020quantum,berberich2024training,du2021quantum,weber2021optimal}}}& Lipschitz Continuity  & Trainable encodings are crucial for adapting robustness by influencing Lipschitz bounds.\\
        \cmidrule{2-3}
        & QDP & Independent of classifier details. Can stem from Depolarization.\\
        \cmidrule{2-3}
        & QHT & Providing robustness conditions in terms of fidelity, Bures metric, and trace distance.\\
        \midrule
        \multirow{3}{*}{\makecell[l]{QPAC Learning\\ \cite{salmon2024provable,franco2024predominant,dowling2024adversarial}}}& Advantages & Achieving a square root improvement in sample complexity over classical methods.\\
        \cmidrule{2-3}
        & Strategies & Unitarity of quantum circuits, dynamical complexity, and data encoding.\\
        \cmidrule{2-3}
        & Scrambling and Chaos &  Making it impossible for adversaries to manipulate the system for misclassification.\\
        \cmidrule{2-3}
        & Resilience & Overparametrized QML models offer inherent protection against gradient inversion attacks.\\
        
        \bottomrule
    \end{tabularx}
\end{table*} 
\section{\textbf{Benchmark Datasets and Evaluation Protocols}}
\label{sec:benchmark-datasets-evaluation}
The rigorous evaluation of QAML models is a cornerstone for the field's advancement. Establishing trust and verifying the efficacy of novel attacks and defense mechanisms is critically dependent on standardized and comprehensive evaluation protocols. Such protocols are built upon three essential pillars: a diverse suite of benchmark datasets that capture a range of real-world and theoretical challenges, a robust set of evaluation metrics capable of quantifying performance and security, and the use of realistic simulation frameworks that reflect the conditions of current and future quantum computers.

\subsection{\textbf{Datasets for QAML}}
The selection of appropriate datasets is fundamental for validating QAML research. The literature demonstrates a tiered approach, utilizing classical, quantum-native, and specialized datasets to test models under various conditions.

\subsubsection{\textbf{Classical Computer Vision Benchmarks}}
A primary category involves the adaptation of well-established classical computer vision benchmarks. Studies often begin with the MNIST-family datasets such as original MNIST \cite{lecun2002gradient} and Fashion-MNIST \cite{xiao2017fashion}, not because they are inherently quantum, but because they provide a standardized and interpretable baseline. These datasets enable researchers to isolate the effects of quantum modeling without the noise of unfamiliar data distributions \cite{lu2020quantum, west2023towards, huang2023enhancing, chu2023qtrojan, fu2025copyqnn, zhou2025watermarking}. The use of these datasets provides common ground for comparing the performance of quantum classifiers against their classical counterparts. To accommodate the qubit limitations of the NISQ devices, these high-dimensional datasets are often adapted through techniques such as classical preprocessing with autoencoders or by focusing on simpler binary classification sub-tasks \cite{kundu2025adversarial, zhou2025watermarking}.

\subsubsection{\textbf{Quantum-Native Datasets}}
To evaluate QAML models in their native domain, researchers use datasets that are directly derived from quantum mechanical systems. These \reflectbox{"}quantum native" datasets are crucial for verifying that security principles hold for problems, where quantum computers are expected to have a natural advantage. Prominent examples include the ground states of the \underline{1D transverse field Ising model} and the \underline{cluster-Ising model}, which are used for the task of classifying quantum phases of matter \cite{lu2020quantum, gong2024randomized, huang2023enhancing}.

\subsubsection{\textbf{Specialized and Synthetic Datasets}}
\begin{figure}[htbp]
\centering
\begin{tikzpicture}
\begin{axis}[
    ybar,
    bar width=20pt,
    ymin=0,
    width=\linewidth,
    height=6cm,
    enlarge x limits=0.25,
    symbolic x coords={Classical Vision,Specialized,Quantum Native},
    xtick=data,
    xticklabel style={align=center, font=\small},
    yticklabel style={font=\small},
    xlabel={types of datasets},
    ylabel={\# of research works},
    xlabel style={font=\small},
    ylabel style={font=\small},
    ytick={0,5,10,15,20,25,30}
]

\addplot[fill=violet!60] coordinates {(Classical Vision,27) (Specialized,6) (Quantum Native,5)};

\end{axis}
\end{tikzpicture}
\caption{The number of works conducting experiments on each types of data.}
\label{fig:datasets-number-of-works}
\end{figure}
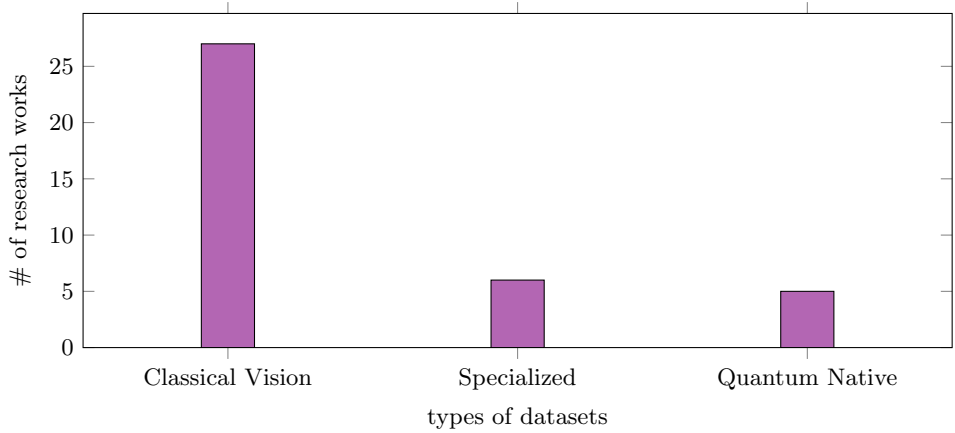

Finally, a growing body of work uses specialized or synthetic datasets to explore the applicability of QAML to specific, security-critical domains. For example, Akter et al. \cite{akter2023exploring} pioneered the use of the \underline{ClaMP malware dataset} to benchmark QML's performance in software supply chain security. For controlled and theoretical experiments, simple synthetic datasets like the \underline{\reflectbox{'}moons' dataset} are used to validate defense mechanisms under ideal conditions \cite{huang2023enhancing}, while sequential data like \underline{\reflectbox{'}sin' functions} have been used to test the vulnerability of quantum recurrent models \cite{chu2023qtrojan}. Figure \ref{fig:datasets-number-of-works} illustrates the number of times that each type of dataset is benchmarked in experimental research works. This shows the extensive usage of the classical datasets compared to the other types, which highlights the need for more experiments on the other types of data. 

\subsection{\textbf{Evaluation Metrics}}
The metrics used to evaluate QAML systems are as diverse as the threats they face, ranging from standard performance measures to highly specialized theoretical bounds. A recurring theme is the necessity of measuring the inherent trade-off between a model's utility and its security.
\subsubsection{\textbf{Core Performance Metrics}}
The most fundamental evaluation involves a pair of core performance metrics: \underline{s}tandard \underline{a}ccuracy (SA), which measures performance on original unperturbed data, and \underline{r}obust \underline{a}ccuracy (RA), which measures the performance on adversarial examples. The trade-off between these two is a central challenge, as many defenses that increase robust accuracy come at the cost of clean accuracy drop \cite{huang2023enhancing}. These are often supplemented by other standard classification metrics such as  precision, recall, F1 score, and loss, which provide a more nuanced view of the behavior of a classifier, particularly when compared with classical models \cite{akter2023exploring, yocam2024quantum}.

\subsubsection{\textbf{Threat-Specific Metrics}}
As the field matures, threat-specific metrics have been developed to evaluate different attack classes. For backdoor attacks, the goal is to simultaneously maximize SA to ensure stealth and the \underline{a}ttack \underline{s}uccess \underline{r}ate (ASR) to ensure effectiveness \cite{chu2023qtrojan}. For model extraction attacks, the evaluation goes beyond accuracy to include efficiency and stealth metrics, such as the total number of queries sent to a victim service and the associated financial cost \cite{fu2025copyqnn}. Similarly, IP protection watermarking techniques are judged by their ability to maintain high accuracy in the primary task while achieving near-perfect accuracy on the hidden trigger set \cite{zhou2025watermarking}.

\subsubsection{\textbf{Formal and Theoretical Metrics}}
Beyond empirical metrics, a significant trend is the development of formal and theoretical measures aiming to provide provable bounds under specific conditions. Research into certified robustness introduces metrics from quantum information theory, such as fidelity, trace distance, and Bures distance between quantum states, to derive mathematical bounds on a model's vulnerability \cite{weber2021optimal}. This theoretical approach is complemented by concepts like QDP, which provides a formal framework for measuring and bounding adversarial risks \cite{gong2024randomized}. Moreover, the intrinsic geometric properties of QML models can be directly linked to their natural adversarial robustness. For instance, in quantum kernel methods, wider SVM margins limit the adversarial perturbation size of the attacker \cite{montalbano2025quantum,lu2020quantum}.

\subsection{\textbf{Simulation Frameworks and Platforms}}
The implementation and validation of QAML research are based on a rapidly evolving ecosystem of software and hardware platforms. The choice of framework is critical for the design, simulation, and deployment of quantum circuits.

Most contemporary QAML research is conducted using open source high-level software frameworks. \underline{Qiskit} and \underline{PennyLane} are the most prominently cited platforms, providing the tools to construct and train variational quantum circuits \cite{montalbano2025quantum, chu2023qtrojan, zhou2025watermarking, akter2023exploring, huang2023enhancing}. These frameworks are often used in a hybrid tool chain, integrating with classical libraries like TensorFlow for pre-processing or optimization tasks \cite{akter2023exploring}.

A crucial trend in evaluation is the move beyond idealized, noiseless simulations towards protocols that account for the realities of the NISQ era. This is achieved in two primary ways. First, researchers increasingly test their models using realistic noise models derived from actual quantum hardware, such as the `FakeAlmaden` noise model in Qiskit, to assess performance under practical noise conditions \cite{chu2023qtrojan, kundu2025adversarial}. Secondly, and more significantly, studies are beginning to validate findings by executing experiments directly on cloud-accessible quantum computers, such as the IBM\_Brisbane, IBM\_Kyoto, and IBM\_Quebec processors \cite{montalbano2025quantum, fu2025copyqnn}. This practice is essential to ensure the reproducibility and real-world relevance of QAML research, directly addressing a key challenge in the field.
\section{\textbf{Emerging Trends and Applications}}
\label{sec:emerging-trends-and-applications}
As the foundational vulnerabilities of QML become clearer, the QAML field is rapidly evolving beyond initial proofs of concept. This evolution is characterized by an \reflectbox{"}arms race" of increasingly sophisticated attack and defense methodologies, a shift toward a more holistic, systems-level view of security, and the application of these principles to new, security-critical domains. This section reviews these emerging trends and applications, highlighting the maturation of the field from theoretical concerns to practical security engineering.

\subsection{\textbf{The Maturation of the Threat Landscape}}

Early research in QAML focused mainly on adapting classical evasion attacks, demonstrating that quantum classifiers, like their classical counterparts, are vulnerable to small input perturbations \cite{lu2020quantum}. However, a significant emerging trend is the exploration of more advanced threats that target the entire machine learning lifecycle, compromising model integrity, confidentiality, and intellectual property.

The threat model has expanded from simple evasion attacks to a broader range of integrity and confidentiality threats, many of which are quantum analogues to well-established vulnerabilities in classical deep learning \cite{liu2021privacy}. Researchers have demonstrated, for example, that \underline{data poisoning}, a well-known classical vulnerability, poses a serious threat to QML systems. Kundu and Ghosh \cite{kundu2025adversarial} introduced a quantum noise-resistant poisoning attack, which intelligently flips training labels to cause maximal confusion and degrade the model accuracy by more than 90\%. This work signals a shift from attacking the final model to corrupting the training process itself.

Furthermore, recent studies have discovered novel attack surfaces unique to the quantum domain that go beyond data manipulation. A prime example is the development of \underline{circuit-level backdoors}, such as the QTrojan attack proposed by Chu et al. \cite{chu2023qtrojan}. In this paradigm, the adversary embeds a Trojan not by poisoning data, but by inserting a few malicious quantum gates directly into the model's circuit architecture. This in theory can be activated, representing a potent supply-chain vulnerability. Currently, the focus has also been broadened to include threats against the model as a commercial asset. The rise of Quantum-as-a-Service (QaaS) platforms has motivated research into \underline{model extraction and IP theft attacks}, where an adversary's goal is not to fool the model, but to steal its intellectual property. CopyQNN \cite{fu2025copyqnn} demonstrates a practical method by using advanced transfer learning techniques to reconstruct a high-fidelity copy of a proprietary model. It has been reported to require 90 times fewer queries than previous methods, highlighting serious economic and security risks in a cloud-based quantum ecosystem.
\subsection{\textbf{The Evolution of Defense Strategies}}

In response to this expanding threat landscape, defense strategies are evolving from direct adaptations of classical methods to more sophisticated, quantum-native, and provable solutions. Initial works focused on applying proven techniques like adversarial training to the quantum realm, showing its effectiveness for both QNNs and quantum kernel methods \cite{lu2020quantum, montalbano2025quantum}.

However, recognizing the limitations of these first-generation defenses, such as their high computational overhead and negative impact on standard accuracy, has led to the development of more advanced, learnable defense mechanisms. For instance, Huang and Zhang \cite{huang2023enhancing} proposed the integration of trainable noise layers into a QNN, allowing the model to learn the optimal level of defensive noise, thereby mitigating the traditional trade-off between robustness and accuracy. Another clever approach repurposes concepts from quantum computing itself, such as using the phenomenon of barren plateaus, typically an obstacle to training, as a defense mechanism to prevent attackers' gradient-based optimization from succeeding \cite{gong2024randomized}.

Perhaps the most significant trend is the shift from purely empirical defenses to those that offer provable security guarantees. This pursuit of certified robustness aims to provide mathematical assurances of a model's safety. Weber et al. \cite{weber2021optimal} established a foundational connection between model robustness and \underline{QHT}, deriving tight, optimal bounds on a classifier's resilience. This approach, rooted in fundamental quantum information theory, which represents a move towards quantum-native security principles. The definition of defense is also expanding beyond attack prevention to include ownership verification. The pioneering work on \underline{watermarking QNNs} in \cite{zhou2025watermarking}, demonstrates how a hidden signature can be embedded within a model to serve as a proof of ownership, addressing the threat of model theft by providing a mechanism for IP dispute resolution, especially relevant in scenarios involving cloud-based quantum services. The relationship between threats and their corresponding defensive paradigms is summarized in Table \ref{tab:attack_defense}.
\begin{table*}[h]
    \centering
    \caption{Effectiveness of the proposed defense strategies against adversarial threats. Full circles (\CIRCLE) denote high empirical or theoretical effectiveness; half-filled circles (\LEFTcircle) indicate partial protection; and empty circles (\Circle) represent unproven or ineffective defenses under current experimental works - adversarial training (AT), gradient masking (GM), randomized encoding (RE), quantum error correction (QEC), quantum physical unclonable function (QPUF), logic locking (LL), software stack exploitation (SSE), and watermarking (WM).}
    \resizebox{\linewidth}{!}{
    \begin{tabular}{l | c c c c c c c c c c c c}
        \hline
        Attack/Defense & AT & Preprocessing & \makecell{GM} & \makecell{Certified} & \makecell{Quantum\\Noise} & \makecell{RE} & QEC & Obfuscation & QPUF & LL & WM\\
        \hline
        Evasion&\CIRCLE&\LEFTcircle&\LEFTcircle&\CIRCLE&\LEFTcircle&\CIRCLE&\Circle&\Circle&\Circle&\Circle&\Circle\\
        Data Poisoning&\Circle&\CIRCLE&\Circle&\LEFTcircle&\LEFTcircle&\LEFTcircle&\LEFTcircle&\Circle&\Circle&\Circle&\Circle\\
        SSE&\LEFTcircle&\Circle&\LEFTcircle&\Circle&\CIRCLE&\LEFTcircle&\Circle&\CIRCLE&\Circle&\LEFTcircle&\Circle\\
        IP Theft&\Circle&\Circle&\Circle&\CIRCLE&\CIRCLE&\CIRCLE&\Circle&\CIRCLE&\CIRCLE&\CIRCLE&\CIRCLE\\
        Fault Injection&\Circle&\Circle&\Circle&\Circle&\Circle&\Circle&\CIRCLE&\Circle&\CIRCLE&\CIRCLE&\Circle\\
        Hardware Specific&\Circle&\Circle&\Circle&\Circle&\CIRCLE&\LEFTcircle&\LEFTcircle&\CIRCLE&\CIRCLE&\CIRCLE&\Circle\\
        \hline
    \end{tabular}}
    \label{tab:attack_defense}
\end{table*}
\subsection{\textbf{A Holistic Approach to Security and New Applications}}
\begin{figure*}[b]
    \centering
    \includegraphics[width=\linewidth]{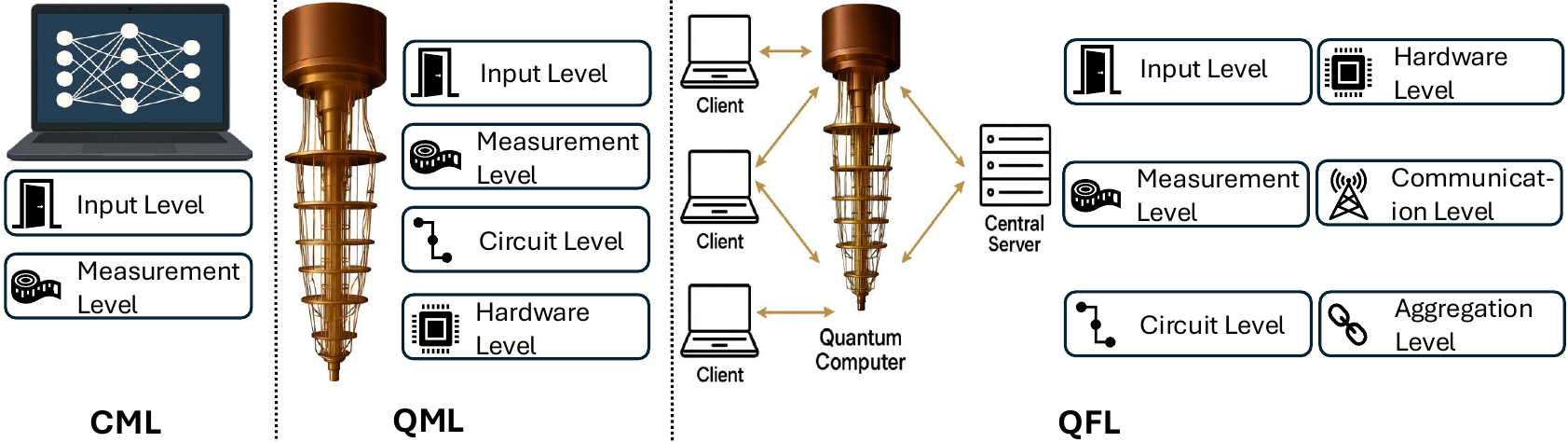}
    \caption{Critical properties that affect the attack surface in CML, QML, and QFL.}
    \label{fig:attack-surface}
\end{figure*}
Beyond the algorithmic arms race, a broader trend in QAML security is the call to treat it as a system-level engineering challenge. High-level reviews advocate for a \reflectbox{"}Security by Design" philosophy, urging the community to learn from the history of classical software by integrating security into the entire development lifecycle \cite{arias2023let, yocam2024quantum}. This perspective, a core tenet of the emerging field of \underline{q}uantum \underline{s}oftware \underline{e}ngineering (QSE), argues that true trustworthiness will come from mature engineering practices, including robust static analysis tools, formal testing, and verifiable development processes, which are currently lacking \cite{arias2023let}. This view also reframes hardware noise not just as a performance issue, but as a potential security vector, highlighting the need to secure the hardware-software interface against threats like fault injection or the exploitation of physical properties like qubit crosstalk \cite{franco2024predominant}.

These emerging security principles are being explored in the context of new and critical applications. Beyond foundational tasks in scientific discovery, such as classifying phases of matter \cite{lu2020quantum}, QAML is being benchmarked for real-world cybersecurity tasks like malware detection \cite{akter2023exploring}. Furthermore, there is a growing recognition that robustness is essential for future distributed learning paradigms.

Researchers have identified the need to extend security guarantees to \underline{q}uantum \underline{f}ederated \underline{l}earning (QFL) architectures, where multiple clients collaborate to train a model without sharing private data. In QFL, either the server, the clients, or both could be quantum computers \cite{gong2024randomized,kannan2024quantum}. \underline{Q}uantum \underline{f}ederated \underline{a}dversarial \underline{l}earning (QFAL) investigates the vulnerabilities of QFL models and aims to design proper defense systems. Attack surface on this area includes all the QAML's attack surface plus communication and aggregation level attack surface as exclusive ones to federated settings. Aggregation level surface arises from the integration of the gradients, where malicious clients or server can affect the global model or where an adversary can exploit the vulnerabilities of the aggregation algorithm. Communication level attack surface, on the other hand, stems from the frequent exchanges between the clients and the  server, where an adversary can include attacks such as eavesdropping, traffic manipulation, and many other attack types \cite{9746622,nguyen2025quantumfederatedlearningcomprehensive,maouaki2025qfalquantumfederatedadversarial,qi2024model,qiao2024transitioning}. Figure \ref{fig:attack-surface} shows how the attack surface on CML, QML, and QFL differs. This intersects with the application of formal privacy-preserving methods, such as QDP, which are being investigated not only for data protection and may also contribute to adversarial robustness, though this remains an open research question \cite{gong2024randomized, franco2024predominant}. A summary of these key application domains is presented in Table \ref{tab:applications}.

\begin{table*}[h]
\centering
\caption{A summary of emerging application domains for QML and QAML.}
\label{tab:applications}
\begin{tabularx}{\linewidth}{@{} p{0.4\columnwidth} >{\RaggedRight}X >{\RaggedRight}X @{}}
\toprule
\textbf{Application Domains} & \textbf{Key Task(s)}&\textbf{Real-World Application(s)}
\\
\midrule
\textbf{Scientific Discovery} \cite{lu2020quantum} & Integrity of QML models for physics research. & Classifying quantum phases of matter.\\
\hline
\textbf{Cybersecurity} \cite{akter2023exploring, chu2023qtrojan}& Security-critical tasks. & Malware detection, Defense development.\\
\hline
\textbf{Distributed Learning} \cite{gong2024randomized} & Robustness guarantees to privacy-preserving paradigms. & QFL.\\
\hline
\textbf{Commercial IP Protection} \cite{fu2025copyqnn, zhou2025watermarking} & Ownership verification methods.&  Watermarking.\\
\bottomrule
\end{tabularx}
\end{table*}

{From a broader perspective, the geographical distribution of QAML research appears to be closely tied to the availability of quantum computing infrastructure and software ecosystems. Many experimental studies rely on commonly used simulation platforms such as Qiskit, Pennylane, and other quantum simulation frameworks, which are primarily developed and supported by institutions and organizations in North America and Europe, especially the United Kingdom. This suggests that current research activity in QAML appears to be concentrated in regions possessing established quantum computing ecosystems and access to cutting-edge computational resources. However, the increasing accessibility of cloud-based quantum platforms is leading to broader global participation, and the field is expected to become more geographically diverse as quantum technologies advance.}
\section{\textbf{Challenges and Open Problems}}
\label{sec:challenges-and-open-problems}
Despite the rapid progress in QAML, the practical deployment of secure and reliable models is fraught with significant hurdles. These challenges span from the physical limitations of current hardware to deep theoretical questions about the nature of quantum learning itself. This section surveys the landscape of these open problems, focusing on the key areas of scalability, robustness in noisy environments, the need for better benchmarks, and the overarching quest for explainable and trustworthy quantum AI.
\subsection{\textbf{Scalability and the NISQ-era Resource Bottleneck}}
A primary barrier to develop and deploy robust QML systems is the inherent resource limitation of the NISQ era. The challenges of scalability manifest in several critical ways. First, the computational overhead of many proposed defense mechanisms is substantial. Adversarial training, while effective in improving robustness against certain input perturbations, imposes a significant computational cost, with research indicating that it can require between 3 to 30 times more training time than a standard, non-robust model \cite{huang2023enhancing}. This overhead is a significant impediment on already scarce and costly quantum computational resources.

Second, the scalability of security analysis itself is a major open problem. Many powerful techniques for crafting attacks or verifying defenses are computationally intractable for all but the smallest systems. For example, the QUID poisoning attack relies on calculating the density matrix of quantum states, a process whose cost scales exponentially with the number of qubits, making it impractical for larger models \cite{kundu2025adversarial}. A promising research direction to mitigate this is the development of efficient classical machine learning models that can approximate these quantum properties, thereby making security analysis feasible for more complex systems.

{Third, most of the existing QAML studies are conducted on small-scale quantum circuits and shallow circuit depths. Although such settings are necessary due to current hardware limitations, they raise concerns about the generalizability of the reported results. Many adversarial effects and defense mechanisms in small circuits may not scale linearly to larger quantum models, where phenomena such as barren plateaus, noise accumulation, and increased parameter interactions become more noticeable. As a result, robustness claims from small-scale experiments should be interpreted carefully, since they may not accurately represent the behavior of QML systems in realistic, large-scale deployment scenarios.}
\subsection{\textbf{Robustness Under Quantum Noise and Decoherence}}
The inherent noise and decoherence of NISQ-era devices represent a fundamental security challenge. This problem is twofold: noise degrades a model's legitimate performance, but it can also be used as a defense.

High-level reviews in the field categorize reliable noise handling as a key pillar of QAML security \cite{west2023towards}. Beyond generic decoherence, a sophisticated threat model includes an adversary who actively manipulates the physical hardware. This could involve targeted \underline{fault injection attacks} to induce misclassifications or the exploitation of device-specific properties, such as \underline{qubit crosstalk} in superconducting systems, to create new attack vectors not present in classical systems \cite{franco2024predominant,upadhyay2024stealthy}. This reframes the hardware-software interface as a critical security boundary, where physical imperfections become potential vulnerabilities \cite{arias2023let}.

This \reflectbox{"}dual-use" nature of noise, where it can be both a vulnerability and a potential defense, creates a complex optimization problem. The development of defenses that are themselves robust to the underlying noise of the system remains a significant open area of research.
\subsection{\textbf{Lack of Standard Benchmarks}}
{The absence of standardized datasets poses a significant challenge in QAML. Current research studies frequently use heterogeneous datasets, diverse encoding strategies, and inconsistent evaluation protocols, which prevent fair and reproducible comparisons across different studies. For instance, some studies evaluated robustness using downscaled classical datasets (e.g., binarized versions) with simplified quantum encodings, while others used synthetic datasets adapted to particular quantum circuits. Moreover, perturbation constraints and attack configurations vary widely across studies. Performance metrics (e.g., robust accuracy and attack success rate) also lack consistency across studies. As a result, reported improvements are usually unreliable. Handling this challenge requires the systematic creation of standardized and hardware-aware benchmarks that clearly specify datasets, circuit configurations, threat models, and evaluation metrics. Such standardization would improve reproducibility and facilitate the advancement of QAML from small prototype demonstrations to practical and real-world implementations.}
\subsection{\textbf{Reproducibility in NISQ Settings}}
{A critical challenge is to bridge the gap between the results obtained from noiseless simulations and performance on real NISQ hardware. The difficulty of reproducing results across different, noisy quantum processors is a well-known problem in the broader quantum computing field. For QAML, this is particularly acute \cite{gong2024randomized}. Without such empirical validation, the true practical utility of many proposed security measures remains uncertain. This also raises concerns about whether current experimental designs, particularly those relying on idealized noise models, are sufficient for evaluating the true robustness of QML systems in realistic deployment settings.

Another important limitation in current QAML research is the absence of statistical completeness in the experiments. Many studies rely on a single or limited experimental run to report robustness improvements, without considering randomness arising from quantum noise, stochastic training procedures, or hardware-specific variations. As a result, the reported robustness gains may not be consistent across multiple executions or different quantum simulation platforms. This concern closely aligns with reproducibility challenges in NISQ settings, where measurement noise and device variability can significantly affect outcomes. To address this issue, future QAML studies should adopt more rigorous evaluation protocols, including multiple experiments and  reporting confidence intervals. These practices ensure the reliability, reproducibility, and comparability of results in the QAML domain. In particular, reporting confidence intervals and analyzing the stability of attack success rates across multiple runs are necessary for evaluating robustness in NISQ settings.}
\subsection{\textbf{Quantum Explainability and Trustworthiness}}
The most profound and long-term challenge facing QAML is the quest for models that are not only empirically robust but fundamentally explainable and trustworthy. This goes beyond defending against attacks to understanding the root causes of vulnerabilities and building systems with provable guarantees.

A central open question is \textit{why} QML models are vulnerable in the first place. Research suggests that this vulnerability is not a superficial flaw, but may be linked to deep, intrinsic properties of high-dimensional spaces. This includes the phenomenon of concentration of measures, where data points in a high-dimensional Hilbert space can become difficult to distinguish, making the model susceptible to small perturbations \cite{west2023towards, montalbano2025quantum}. Some have even suggested a link to fundamental physics, suggesting a connection between adversarial vulnerabilities and the orthogonality catastrophe \cite{lu2020quantum}. Understanding these root causes is a critical step towards building inherently robust models.

This lack of explainability is exacerbated by a discipline-wide challenge: the immaturity of QSE practices. The absence of robust tools such as debuggers, static analyzers, and standardized testing libraries makes it incredibly difficult to verify and validate the behavior of complex quantum software \cite{arias2023let}. This \reflectbox{"}inscrutability" of quantum circuits creates a perfect hiding place for sophisticated threats like circuit-level backdoors, directly linking the challenge of explainability to a concrete security risk \cite{chu2023qtrojan}.

Ultimately, trustworthiness is also a legal concern. As QML models become valuable IPs, protecting them from theft is of paramount importance. The demonstration of practical \underline{model extraction} attacks \cite{fu2025copyqnn} and the development of defensive \underline{watermarking} techniques \cite{zhou2025watermarking,roy2025watermarking} highlight that building trustworthy quantum AI involves not only algorithmic security but also the creation of a robust framework for IP protection. The overarching goal is to design a mature engineering discipline capable of producing quantum systems that are secure by design.
\subsection{\textbf{Realistic Threat Models}}
{Another critical gap in QAML research is the difference between theoretical threat models and practical deployment scenarios. Most existing studies assume strong white-box access, where the adversary has full knowledge of the model’s parameters and gradients. However, real-world QML systems are typically accessed through cloud interfaces, making black-box or query-based attacks more realistic. This mismatch limits the practicality of many proposed attacks, highlighting the need for evaluation under realistic threat models.}
\subsection{\textbf{Lack of Certified Robustness under Noise}}
{Another gap is the lack of certified robustness guarantees that remain valid under realistic quantum noise and measurement uncertainty. While several theoretical frameworks (e.g., quantum DP and Lipschitz-based bounds) provide formal guarantees, these often rely on ideal assumptions. In practical NISQ devices, hardware noise, decoherence, and measurement variability can significantly weaken these guarantees, raising concerns about their applicability in real-world scenarios.}

{Overall, these challenges can be distilled into a set of key bottlenecks for QAML: (i) the lack of standardized and hardware-aware evaluation benchmarks, (ii) the mismatch between theoretical threat models and practical deployment scenarios, (iii) the tension between robustness and resource constraints in NISQ devices, (iv) the absence of certified robustness guarantees under realistic quantum noise, and (v) the limited reproducibility of results across simulators and real quantum hardware. Addressing these bottlenecks is essential for advancing QAML from proof-of-concept studies toward practical and secure quantum machine learning systems.}
\section{\textbf{Conclusion}}
\label{sec:conclusion}
QAML is emerging as a crucial field at the intersection of quantum computing, AI security, and systems engineering. Although early efforts focused on direct adaptations of classical adversarial techniques, the field is now evolving into a discipline of its own, complete with quantum native attacks, formal defenses, and application-specific requirements. A recurring insight is that QAML security cannot be treated as an afterthought; it must be integrated into the entire pipeline, from hardware-aware compilation to federated training and IP protection. Moreover, the increasing realism of simulation platforms and the availability of cloud-accessible quantum hardware will likely accelerate both the discovery of new vulnerabilities and the empirical validation of robust defenses.

{In addition to recent advances, it is necessary to provide practical guidance for future QAML research. Researchers are recommended to use standardized benchmarks and clearly defined threat models when evaluating adversarial robustness. Experimental results should be reported with enough detail, including circuit configurations, datasets, and platform specifications, to ensure reproducibility. Additionally, the researchers need to account for variability in NISQ devices by including multiple runs and reporting performance statistics, such as confidence intervals. Ultimately, evaluating robustness under realistic noise and across different platforms is necessary to understand the reliability of QAML methods. These practices can help establish more consistent, transparent, fair, and comparable evaluation protocols in the field.}

Looking ahead, the field must address two interdependent goals: scalability and explainability. As quantum models grow in complexity and industry, ensuring their trustworthiness becomes crucial. The next generation of QAML research must bridge theoretical insights with rigorous engineering discipline, aligning provable robustness with practical deployability on noisy quantum systems, building systems that are not just clever but verifiably secure, interpretable, and resilient.
\section*{Declarations}
\subsection*{Funding}
This work is partially supported by the Natural Sciences and Engineering Research Council of Canada (NSERC) under funding reference number RGPIN-2021-02968.
\subsection*{Conflict of interest}
The authors declare no conflict of interests.
\subsection*{Author contribution}
R.R.F.: Conceptualization, Formal Analysis, Writing - Original Draft, Reviewing, Editing, Supervision, and Funding Acquisition; M.M.: Validation, Formal Analysis, Writing - Original Draft, Reviewing, and Editing; E.M.: Writing - Original Draft; D.V.: Writing - Original Draft; P.P.: Writing - Original Draft; F.G.: Writing - Original Draft; S.S.: Writing - Original Draft; S.N.: Writing - Original Draft; K.H.: Writing - Original Draft. All authors reviewed the manuscript.
\bibliography{main}

\begin{appendices}
\section{Literature Search and Selection Methodology}
{To ensure transparency, rigor, and reproducibility, this survey follows the PRISMA (Preferred Reporting Items for Systematic Reviews and Meta-Analyses) framework to guide the identification, screening, and selection of relevant studies in QAML.

\textbf{Search Strategy:}
A structured keyword-based search was conducted across a diverse set of academic databases and publication platforms spanning machine learning, quantum computing, and physics communities. These included \textit{IEEE Xplore}, \textit{ACM Digital Library}, \textit{Scopus}, \textit{Web of Science}, \textit{arXiv}, \textit{ScienceDirect}, \textit{SpringerLink}, and major physics-focused venues such as the \textit{Physical Review family of journals (APS)}, including \textit{Physical Review A}, \textit{Physical Review Letters}, and \textit{Physical Review Research}. In addition, relevant studies were identified from interdisciplinary venues and high-impact journals and conferences in machine learning and quantum computing, ensuring broad coverage of both domain-specific and cross-disciplinary contributions in QAML.

The search process was designed around two core concepts central to this survey: adversarial machine learning and quantum machine learning. These anchor terms were systematically combined with a comprehensive set of related keywords to ensure broad and inclusive coverage of the literature.

Specifically, the search queries were constructed using representative keywords and their variations, such as:
\reflectbox{"}adversarial", \reflectbox{"}adversarial attack", \reflectbox{"}adversarial defense", \reflectbox{"}robustness", \reflectbox{"}security", \reflectbox{"}privacy", \reflectbox{"}evasion attack", \reflectbox{"}poisoning attack", \reflectbox{"}model extraction", \reflectbox{"}membership inference", \reflectbox{"}quantum machine learning", \reflectbox{"}quantum adversarial machine learning", \reflectbox{"}quantum neural network", \reflectbox{"}variational quantum circuit", \reflectbox{"}parameterized quantum circuit", \reflectbox{"}quantum classifier", \reflectbox{"}quantum adversarial", \reflectbox{"}quantum security", and \reflectbox{"}quantum robustness". These keywords were not treated as an exhaustive list, but rather as representative terms capturing the main concepts and variations used in the literature.

To maximize recall, the search was applied to both the titles and abstracts of publications, ensuring that relevant studies were captured even if specific keywords were not explicitly present in the title.

\textbf{Selection Process:}
Following the PRISMA methodology, the literature selection was conducted through four stages: identification, screening, eligibility, and inclusion. In the identification stage, candidate studies were retrieved using the defined keyword queries across all selected databases. During the screening stage, titles and abstracts were reviewed to remove clearly irrelevant works and duplicates. In the eligibility stage, full-text articles were assessed based on their relevance to adversarial threats and defenses in QML. Finally, only studies that met the inclusion criteria and provided sufficient technical or experimental detail were retained in the final set of reviewed papers.

\textbf{Inclusion and Exclusion Criteria:}
Studies were included if they (i) addressed adversarial attacks or defenses in quantum machine learning, (ii) proposed, analyzed, or experimentally evaluated QAML methods, and (iii) provided sufficient methodological detail to support technical understanding and reproducibility. Studies were excluded if they (i) focused solely on classical adversarial machine learning without quantum relevance, (ii) lacked technical depth, or (iii) were duplicate or preliminary versions of already considered works.

This systematic and structured approach ensures that the survey is comprehensive, unbiased, and reproducible, and provides a solid foundation for analyzing the current landscape of QAML research.}
\end{appendices}

\end{document}